\documentclass[envcountsame]{llncs}

\usepackage{ellipsis, ragged2e} % bugfixing
\usepackage[l2tabu, orthodox]{nag} % avoid typical LaTeX errors

\usepackage[english]{babel}
\usepackage[T1]{fontenc}
\usepackage{lmodern}
\usepackage[utf8]{inputenc}
\usepackage[final]{microtype}

\usepackage[ruled,noline]{algorithm2e}

\usepackage{bbm}
\usepackage{proof}
\usepackage{csquotes}
\usepackage{mathtools}
\usepackage{booktabs}
\usepackage{multirow}
\usepackage{afterpage}
\usepackage{xparse}
\usepackage{hyperref}
\usepackage{enumitem}

\usepackage[capitalise]{cleveref}

\input{0_macros}
\usepackage{tikz}
\usetikzlibrary{automata,positioning,shapes,calc,fit,arrows.meta,backgrounds}
\usepackage{pgfplots}
\usepgfplotslibrary{fillbetween}
\pgfplotsset{compat=1.13}

\tikzstyle{state}+=[minimum size=20pt,inner sep=2pt]
\tikzstyle{action}=[font=\small,inner sep=0pt,outer sep=3pt]
\tikzstyle{actionedge}=[->,draw]
\tikzset{chainarrow/.tip={Stealth[length=3pt]}}
\tikzset{>=chainarrow}

\tikzstyle{system} = [draw,circle,minimum size=1cm]
\tikzstyle{environment} = [draw,diamond,minimum size=1cm]

\tikzstyle{systemltl} = [draw,rectangle,rounded corners=5pt,minimum size=1cm]
\tikzstyle{environmentltl} = [draw,rectangle,minimum size=1cm]

\tikzset{every picture/.append style={
	auto,
	node distance=2cm,
	initial text={},
}}

\pgfplotscreateplotcyclelist{no marks}{%
	loosely dotted, mark=\empty\\%QL win
	densely dotted, mark=\empty\\%QL prio
	solid, mark=\empty\\%QL sem
	densely dashed, mark=\empty\\%SI
	loosely dashed, mark=\empty\\%SI sem
	dashdotted, mark=\empty\\%QL eps
	dashdotdotted, mark=\empty\\%SI eps
	densely dashdotted, mark=\empty\\%
}

\title{
	Guessing Winning Policies in LTL Synthesis\\
{by Semantic Learning }
\thanks{
	This research was funded in part by the German Research Foundation (DFG) project 427755713 \emph{Group-By Objectives in Probabilistic Verification (GOPro)}
}
}
\author{
	Jan~K\v{r}et{\'i}nsk{\'y}\inst{1,2}\orcidID{0000-0002-8122-2881}
	\and Tobias~Meggendorfer\inst{3,1}\orcidID{0000-0002-1712-2165}
	\and Maximilian~Prokop\inst{1,2}\orcidID{0009-0008-6512-8693}
	\and Sabine~Rieder\inst{1}\orcidID{0009-0006-6397-3100}
}

\institute{
	Technical University of Munich, Germany
	\and
	Masaryk University, Brno, Czech Republic
	\and
	Institute of Science and Technology, Klosterneuburg, Austria
}

\newtoggle{arxiv}
\toggletrue{arxiv}
\usepackage{todonotes}
\begin{document}

\pagestyle{plain}
\maketitle
\begin{abstract}
%We present a new, learning-based approach to solving parity games resulting from LTL synthesis queries.
%We exploit recent advances in the automata-based approach, providing semantic labelling of the vertices in the parity game.
%We utilize this labelling to pick an initial strategy for strategy improvement and extend the idea to a Q-learning-based approach.
%Experimental data suggest the applicability of this idea, encouraging further work to use semantic labels.

We provide a learning-based technique for guessing a winning strategy in a parity game originating from an LTL synthesis problem. A cheaply obtained guess can be useful in several applications. Not only can the guessed strategy be applied as best-effort in cases where the game's huge size prohibits rigorous approaches, but it can also increase the scalability of rigorous LTL synthesis in several ways. Firstly, checking whether a guessed strategy is winning is easier than constructing one. Secondly, even if the guess is wrong in some places, it can be fixed by strategy iteration faster than constructing one from scratch. Thirdly, the guess can be used in on-the-fly approaches to prioritize exploration in the most fruitful directions.

In contrast to previous works, we (i)~reflect the highly structured logical information in game's states, the so-called semantic labelling, coming from the recent LTL-to-automata translations, and (ii)~learn to reflect it properly by learning from previously solved games, bringing the solving process closer to human-like reasoning.

%We exploit semantic labelling.
\end{abstract}
\section{Introduction}

\paragraph*{LTL synthesis} \cite{DBLP:conf/icalp/PnueliR89} is a framework for automatic construction of reactive systems specified by formulae of linear temporal logic (LTL) \cite{Pnueli77}.
Since LTL is a prominent logic in the area of safety-critical and provably reliable dynamic systems, LTL synthesis is a very tempting option to construct such systems since it avoids error-prone manual implementation; instead it is replaced with the need for a complete specification of the system
(which is not trivial either, but in some cases easier).
However, there is also an important computational caveat: the problem of LTL synthesis is 2-EXPTIME complete.
Despite the infeasibility in the worst-case, many heuristics have been designed that can cope with practical problems, as documented by the yearly progress in the synthesis competition SYNTCOMP \cite{DBLP:journals/corr/abs-2206-00251}, which has an LTL track for a number of years.
Yet, many reasonable instances even in the benchmark set of SYNTCOMP still remain practically unsolvable.
In this paper, we aim at \emph{guessing a solution} through a machine-learning model, even for hard cases, thus possibly providing an applicable answer, in a sense, without reading the input formula.
We achieve that by learning from other games and by reflecting \emph{semantic} information, bringing the process closer to human reasoning.

The classic technique for solving LTL synthesis is to
\begin{enumerate}
	\item turn the LTL formula into a deterministic parity automaton (DPA),
	\item turn the DPA (and the partitioning of atomic propositions into system variables and environment variables) into a parity game (PG) between the system and the environment players, and
	\item solve the PG; any winning strategy of the system player then directly induces a system policy (also representable as a circuit) satisfying the LTL formula.
\end{enumerate}
Due to the worst-case doubly-exponential blowup in the first step and the practically bad performance of (Safra's \cite{DBLP:conf/focs/Safra88} and others' \cite{DBLP:conf/lics/Piterman06,DBLP:conf/fossacs/Schewe09}) determinization procedures, this option was rarely used practically until direct, more practical translations were given \cite{DBLP:conf/cav/EsparzaK14,DBLP:journals/jacm/EsparzaKS20}.
The significantly smaller automata \cite{DBLP:conf/atva/KomarkovaK14} have made this approach feasible and, in fact, winning in SYNTCOMP since then.
The approach is implemented in the tool \texttt{Strix} \cite{DBLP:conf/cav/MeyerSL18}, which additionally constructs the DPA/PG \emph{only partially}, on-the-fly until it finds a winning strategy for one of the players.
This helps to overcome some more cases where the DPA is still very large; yet, more complex specifications often remain out of reach.

\paragraph*{Semantic Labelling}
The key difficulty in the on-the-fly exploration %of the PG
is a good heuristic that prioritizes exploration in %the most 
promising directions, so that a solution can be obtained quickly, without constructing \enquote{irrelevant} parts of the game.

\emph{In a concrete state of a PG, is it better to go left or right?}
While this question obviously does not have a simple answer in general, we take a step back and instead of a PG we solve the LTL synthesis problem.
For instance, consider a state of a PG corresponding to satisfying $\ltlGlobally a$, i.e.\ \enquote{always $a$ holds}.
Then, the letter $\{a\}$ is clearly a better choice (for the system) than $\emptyset$.
The former leads to the obligation of satisfying again $\ltlGlobally a$; the latter to the obligation $\false$ (falsifying the formula).
Taking the former edge does not guarantee winning, but the chances are certainly higher than giving up directly.
In order to estimate the chances of winning with some obligation, we can evaluate it by randomly assigning truth values to temporal subformulae; intuitively, $\ltlGlobally a$ can be true or false, so its \enquote{trueness} is 0.5, $\false$ has trueness $0$.
\emph{Trueness} is examined in \cite{DBLP:conf/atva/KretinskyMM19} and utilized in newer versions of \texttt{Strix} \cite{DBLP:journals/acta/LuttenbergerMS20} as guidance.

\emph{Does every state correspond to a goal in LTL? And if so, can we determine which continuation brings us closer to satisfying it?}
Recall that the classic translations of LTL to non-deterministic B\"uchi automata (NBA), stemming from \cite{DBLP:conf/lics/VardiW86}, label the states of the NBA with a conjunction of LTL formulae, which are the current goals in this state.
For deterministic automata, the situation is inevitably more complex.
While the determinization procedures obfuscated any possible such semantic labelling, the more recent approach re-established it, e.g., \cite{DBLP:conf/cav/EsparzaK14} with \cite{DBLP:journals/acta/KretinskyMWW22}, or \cite{Sickert_16_LDBA} with \cite{DBLP:conf/tacas/EsparzaKRS17}.
Beside the overall goal, it is necessary to also monitor the \emph{progress of subgoals}.
For example, consider $\ltlGlobally\ltlFinally (a\wedge\ltlNext b)$ \enquote{infinitely often $a$ is followed by $b$}.
No matter what happens, the goal remains the same.
However, whenever $a$, we are progressing with the subgoal of seeing the $a-b$ sequence once, yielding a subgoal $b$, which is regarded as promising.

\begin{figure}[t]
	\centering
	\begin{tikzpicture}[auto,yscale=0.8]
		\node[systemltl, initial left] (1) at (0,0) {$v_1$};
		\node[systemltl] (2) at (3,1) {$v_2$};
		\node[systemltl] (3) at (3,-1) {$v_3$};
		\node[systemltl] (T) at (6,0) {\text{goal}};
		
		\path[->]
			(1) edge[loop above] (1)
			(1) edge[bend left=15] (2)
			(1) edge[bend right=15,swap] (3)
			(2) edge[bend right=20,swap] (3)
			(3) edge[bend right=20,swap] (2)
			% (3) edge[loop below] (3)
			(2) edge[bend left=15] (T)
			(3) edge[bend right=15,swap] (T)
			(T) edge[loop above] (T)
		;
	\end{tikzpicture}
	\caption{
		Simple game where it is not clear which edges are \enquote{winning}.
	}\label{fig:SI_bad}
\end{figure}
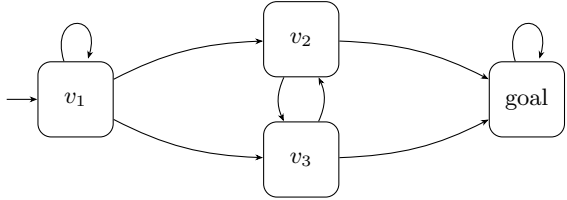

\paragraph*{Our Aim} In this paper, we aim at \emph{better guessing of winning decisions} than in \cite{DBLP:conf/atva/KretinskyMM19,DBLP:journals/acta/LuttenbergerMS20}.
While the previous work only reflected trueness of the main goal, which is just the percentage of truth assignments leading to satisfaction of a Boolean formula, our approach reflects also (i)~the temporal structure of the formulae, (ii)~the monitored subgoals, and (iii)~learns from previously solved games.
On the technical level, we design over 200 \emph{structural features} instead of just trueness, learn an \emph{SVM} classifier comparing which edge is most promising, and use \emph{data from previously solved games}, i.e.\ which edges are \enquote{winning}.
As it turns out, defining this notion already is surprisingly tricky:
We cannot simply use the output of classical strategy improvement algorithms, as there may be multiple, incompatible solutions.
Indeed, already for reachability, there are no maximal permissive strategies \cite{DBLP:journals/ita/BernetJW02}, see \cref{fig:SI_bad}.
Here the edge $(v_2, v_3)$ is winning iff $(v_3, v_2)$ is not used, and vice versa; using both makes them losing.
Nevertheless, they are \enquote{better} than, e.g., the self-loop on $v_1$, which is always losing.
Thus, we want to value both edges between $v_2$ and $v_3$ equally, and higher than the self loop on $v_1$.

%How ``better'' our guessing is can and will be measured in various ways: are the decisions more often optimal? is the guessed strategy closer to a winning one? how often is it winning the whole game? does it provide a better guidance in the on-the-fly exploration?
%Interestingly, it is not even clear how to measure \enquote{how winning} an edge is.

\paragraph*{Our Contribution} can be summarized as follows:
\begin{itemize}
	\item We learn a model predicting which edge has better chances to be winning.
	To this end, we define features on the semantic labelling in \cref{sec:handling:features}, introduce a way to measure the degree of \enquote{winning} of an edge in \cref{sec:ground_truth}, and apply learning of support vector machines using our novel ground truth in \cref{sec:handling:svm}.
	\item We evaluate \enquote{how winning} the suggested strategy is, i.e.\ how many wrong choices it made, on several inputs in \cref{sec:experiments:atva}.
	Surprisingly, this value often is $0$, i.e.\ our strategy is often winning even for complex formulae, and even without reading them (meaning that our strategy is of constant size, \emph{independent of the formula}, as opposed to a decision table in the concrete game; it can be run on the fly with no pre-computation, and decisions depend only on the labelling of the current state).
	\item Besides, while \texttt{Strix}'s architecture and interface ask for a significantly different type of advice (not just for the better of two edges), %and not just for one DPA)
	we show \texttt{Strix} already profits from our advice and---modulo our unoptimized advice implementation---speeds up significantly, as we see in \cref{sec:experiments:strix}.
\end{itemize}

\paragraph*{Usage of our Results:}
\begin{itemize}
	\item We provide an immediate solution (without even reading the input formula), which is often winning; moreover, it is applicable even to games too huge to be analyzed in any way.
	Besides, it is even of a constant size, i.e.\ independent of the size of the state space.
	\item Our approach opens the way to (i)~a solver based on the semantic labelling, for instance, based on strategy iteration only quickly fine-tuning the already almost correct guess, and (ii)~an on-the-fly-exploration advisor to \texttt{Strix}, with the proven potential to be the most efficient among the current techniques.
\end{itemize}

\paragraph*{Related work}
To the best of our knowledge, there is only one other approach to using machine learning in LTL-synthesis.
Here, the authors train a very powerful model (a hierarchical transformer) in order to directly predict a controller or counter example solely off the LTL specification \cite{DBLP:conf/nips/SchmittHRF21}.
Further, if their prediction is refuted by a classical model checking algorithm, they train a separated hierarchical transformer to repair it \cite{DBLP:journals/corr/abs-2303-01158} until it is correct.
While this turns out to be an overall competitive approach that also manages to solve some instances where classical synthesis tools as \texttt{Strix} \cite{DBLP:conf/cav/MeyerSL18} fail, this does not yield a complete procedure, as the repair loop is not guaranteed to ever terminate.
In this work, we aim to improve existing, complete procedures such as implemented in \texttt{Strix} by means of machine learning based heuristics.

\section{Preliminaries}

We introduce notation and provide an overview of necessary background knowledge.
Due to space constraints, we only briefly comment on several topics and refer the interested reader to the respective literature.

We use $\Naturals$ to denote the set of non-negative integers.
%Given a propositional formula $\phi$ over a set of propositions $\AP$, we use $\satisfying(\phi) = \{v \in 2^{\AP} \mid v \models \phi\}$ to denote the set of all satisfying assignments.
The constants $\true$ and $\false$ denote \emph{true} and \emph{false}, respectively.

\subsection{Synthesis \& Games}
The synthesis problem in its general form asks whether a system can be controlled such that it satisfies a given specification under any (possible) environment.
%For example, we can ask whether there exists a mutual exclusion mechanism that provides thread safety under all possible interleavings.
%Here, the system is the mutual exclusion mechanism, the environment is given by the scheduler, and the specification is that no two threads access the critical region simultaneously.
Moreover, one often is interested in obtaining a witness to this query, i.e.\ some \emph{controller} or \emph{strategy} which specifies the system's actions.
%For example, one might ask whether a robot can be steered over difficult terrain such that it arrives at a particular target location.

\begin{figure}[t]
	\centering
	\begin{tikzpicture}[auto,node distance=1.5cm,initial text=]
		\node[systemltl,initial left] (s0) {$v_0, 4$};
		\node[environmentltl] [right of=s0] (s1) {$v_1, 2$};
		\node[systemltl] [below of=s1] (s2) {$v_2, 1$};
		\node[environmentltl] [right of=s1] (s3) {$v_3, 3$};
		\node[systemltl] [below of=s3] (s4) {$v_4, 5$};

		\path[->]
			(s0) edge[loop above] (s0)
			(s0) edge (s1)
			(s0) edge (s2)

			(s1) edge (s3)
			(s1) edge[bend right] (s2)

			(s2) edge[bend right] (s1)
			(s2) edge (s3)

			(s3) edge[loop above] (s3)
			(s3) edge (s4)

			(s4) edge[loop right] (s4)
		;
	\end{tikzpicture}

	\caption{
		An example parity game, taken from \cite{DBLP:conf/atva/KretinskyMM19}.
		Rounded rectangles belong to the system $\sys$ and normal rectangles to the environment $\env$.
		The vertices are additionally labelled with their priorities.
	} \label{fig:example_game}
\end{figure}
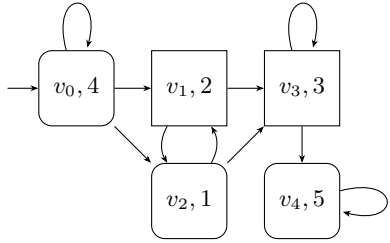

\paragraph{Parity Games} are a standard formalism used in synthesis.
A \emph{parity game} is a tuple $\G = ((\Vertices, \Edges), \initialvertex, \VertexPlayer, \priorities)$, where
	$(\Vertices, \Edges)$ is a finite digraph,
	$\initialvertex \in \Vertices$ a \emph{starting vertex},
	$\VertexPlayer : \Vertices \to \{\sys, \env\}$ a \emph{player mapping}, and
	$\priorities : \Vertices \to \Naturals$ a \emph{priority assignment}.
Each vertex belongs to one of the two players $\sys$ (called \emph{system}) and $\env$ (called \emph{environment}).
In other words, the set of vertices is partitioned into player $\sys$'s vertices $\VerticesSys$ and player $\env$'s vertices $\VerticesEnv$.
See \cref{fig:example_game} for an example.
%For ease of notation, we write $v \Edges := \{(v, u) \in \Edges \mid u \in \Vertices\}$ to denote all outgoing edges of a vertex $v$. % and define $E_i := \{(u, v) \in \Edges \mid u \in \Vertices_i\}$ the set of all edges \enquote{controlled} by player $i$.
%For simplicity, we assume in the following that all games are \emph{alternating}, i.e.\ the successors of a vertex belong to a different player than the vertex itself.

\begin{remark}
	In our implementation priorities are assigned to edges instead of vertices, as this allows for a much more concise representation and suits most translations better.
	However, for ease of presentation, we consider \emph{state-based acceptance} instead of \emph{transition-based}.
	
\end{remark}

\paragraph{Playing}
To play the game, a token is placed in the initial vertex $\initialvertex$.
Then, the player owning the token's current vertex moves the token along an outgoing edge of the current vertex.
This is repeated infinitely, giving rise to an infinite sequence of vertices containing the token $\play = \initialvertex v_1 v_2 \cdots \in \Vertices^\omega$, called a \emph{play}.
%The set of all possible plays is denoted by $\Plays$.
We write $\play_i$ to refer to the $i$-th vertex in a play.
A play $\play$ is \emph{winning} (for the system player) if the smallest priority occurring infinitely often is odd.
(Using \enquote{maximal} instead of \enquote{minimal} or \enquote{even} instead of \enquote{odd} does not fundamentally change the problem at hand.)
Formally, we define $\inf(\play) = \{v \in \Vertices \mid \forall j.~\exists k \geq j.~\play_j = v\}$ as the set of infinitely occurring states.
Since the game graph is finite, this set always is non-empty.
The smallest priority occurring infinitely often is given as $\priorities(\play) = \min \{\priorities(v) \mid v \in \inf(\play)\}$ and system wins the play $\play$ if $\priorities(\play)$ is odd.

\paragraph{Strategies} A strategy of player $\player$ is a mapping $\strategy_\player : V_\player \to E$ assigning to each of $\player$'s vertices an appropriate edge along which the token will be moved, i.e.\ $(v, \strategy_\player(v)) \in E$ for all $v \in V_\player$.\footnote{
	Strategies may be more complex, e.g., by using memory.
	However, \enquote{positional} strategies are sufficient for parity games, thus we omit the general definition.
}
Once both players fix a strategy, the game is fully determined and a unique run is induced.
%We call a pair of strategies $(\strategy_\sys, \strategy_\env)$ winning if the corresponding run is winning.
We call a strategy of system $\strategy_\sys$ \emph{winning} if for \emph{all} strategies of the environment $\strategy_\env$ the induced play is winning, i.e.\ system wins no matter what the environment does.

For example, consider again the game depicted in \cref{fig:example_game}.
Fixing the strategies $\strategy_\sys = \{v_0 \mapsto (v_0, v_2), v_2 \mapsto (v_2, v_3), v_4 \mapsto (v_4, v_4)\}$ and $\strategy_\env = \{v_1 \mapsto (v_1, v_2), v_3 \mapsto (v_3, v_3) \}$ induces the play $v_0 v_2 v_3 v_3 \cdots$.
The set of infinitely often seen priorities equals $\{3\}$, hence the system player wins with these strategies.
Moreover, the strategy $\strategy_0$ is winning, since the play always ends up in either $v_3$ or $v_4$.

\paragraph{Synthesis}
With these notions, we can compactly define the synthesis question:
\emph{Given a parity game $\G$, does there exist a winning strategy for the system player?}
In the example above, $\strategy_0$ is a witness to this question.

This problem is still intensely studied due to its broad applications.
It also is one of the few problems which canonically lie in $\NP \intersection \coNP$ (even in $\mathbf{UP} \intersection \mathbf{coUP}$ \cite{DBLP:journals/ipl/Jurdzinski98}), with recent breakthroughs achieving quasi-polynomial algorithms \cite{DBLP:conf/spin/FearnleyJS0W17,DBLP:journals/siamcomp/CaludeJKLS22,DBLP:journals/lmcs/LehtinenPSW22}.

\paragraph{Extensive-Form Game}
A common notion in game theory is the \emph{extensive-form} game.
Intuitively, this means completely \enquote{unrolling} the game into an explicit representation.
See e.g.\ \cite[Chp.~5-7]{osborne2004introduction} for details.
In our case, we consider the \emph{game tree}, where each node corresponds to a simple path in the game $\G$.
Suppose we are in state $s = (v_1, \dots, v_i)$ of the game tree.
Then, the successors of $s$ are determined by all successors of $v_i$ in the game, i.e.\ $\{u \mid (v_i, u) \in E\}$ as follows.
Suppose such a successor $u$ already occurs along $s$, i.e.\ a loop is closed, we check if the corresponding play is winning or losing.
In that case, the choice leads to a corresponding winning or losing leaf of the tree, respectively.
Otherwise, i.e.\ when no loop is closed by the choice, it leads to $s \circ u$.
Essentially, this game tree represents all potential simple paths (and thus, intuitively, all potential positional strategies) that can arise in the game, and each edge corresponds to a particular move of a player (also called \emph{ply} in game theory).
In particular, it is finite, however of potentially exponential size.
Note that we can restrict to simple paths only because positional strategies are sufficient.

\paragraph{Minimax Game Solving}
A fundamental way to solve games is the \emph{minimax decision} rule, which intuitively corresponds to exhaustively exploring the extensive-form game (also discussed in \cite{osborne2004introduction}).
Suppose we assign a value of $0$ to \enquote{losing} leaves of the game tree and a value of $1$ to the \enquote{winning} leaves.
Then, we can \enquote{back-propagate} values by setting $V(s)$ the maximum of all successors of $s$ if it currently is the turn of the system player and the minimum if instead it is environment's turn (which wants the system to lose).
The game is winning if the value in the initial state of the game tree is $1$.
This approach is also called \emph{backward induction} or \emph{retrograde analysis}: starting from the winning / losing positions of the game, we consider all moves which could lead to such situations.

\paragraph{Strategy Improvement} (or \emph{strategy iteration}, abbreviated by \emph{SI}) is the most prominent practical way of solving parity games, i.e.\ answering the synthesis question.
It received significant attention due to recent practical advances \cite{DBLP:conf/atva/HoffmannL13,DBLP:conf/cav/Fearnley17,DBLP:conf/atva/MeyerL16,DBLP:conf/atva/FriedmannL09} and modern tool developments \cite{DBLP:conf/tacas/Dijk18,DBLP:conf/cav/MeyerSL18}.
We explain the approach briefly, since its details are not important for this work.
Intuitively, SI starts from arbitrary initial strategies for each player, and then performs the following steps in a loop.
First, we check whether either strategy is winning.
If yes, the algorithm exits, returning this strategy.
Otherwise, one of the strategies is improved by changing its choices in some vertices.
If an improvement is not possible, there exists no winning strategy for the respective player.
Otherwise, the process is repeated with the new strategy.

This algorithm converges to the correct result in finite time for any initial strategy.
However, if this initial strategy is chosen \enquote{close} to a winning strategy, then SI intuitively needs to perform fewer steps to converge to an optimal one.
Thus, a heuristic which often comes up with a \enquote{good} initial strategy may improve the runtime significantly over arbitrary or random initialization. %, since then only a few improvement steps are necessary.

%Throughout this work, we refer to a reference implementation of SI, denoted \texttt{SI}, e.g., when running SI with a particular initial strategy.
%In our implementation, we used the algorithm of \cite{DBLP:conf/cav/VogeJ00}, but other variants could be substituted.
%
\subsection{Linear Temporal Logic and Reactive Synthesis}

\paragraph{Linear Temporal Logic} (LTL) \cite{Pnueli77} is a standard logic used to specify desired behaviour of a system.
The syntax usually is given by
\begin{equation*}
	\phi ::= \lfalse \mid a \mid \lnot \phi \mid \phi \land \phi \mid \ltlNext \phi \mid \phi \ltlUntil \phi,
\end{equation*}
where $a \in \AP$ is an \emph{atomic proposition}, inducing the \emph{alphabet} $\Alphabet = 2^\AP$.
These formulae are interpreted over infinite sequences $w \in \Alphabet^\omega$ called $\omega$-words.
A word $w = w_0 w_1 \cdots \in \Alphabet^\omega$ satisfies the \emph{next} operator $\ltlNext \phi$ iff $\phi$ is satisfied in the next step.
Similarly, the \emph{until} operator $\phi \ltlUntil \psi$ is satisfied iff $\phi$ holds until $\psi$ is eventually satisfied.
Usual abbreviations are defined as \emph{finally} $\ltlFinally \phi \equiv \ltrue \ltlUntil \phi$ and \emph{globally} $\ltlGlobally \phi \equiv \lnot \ltlFinally \lnot \phi$, which require that $\phi$ holds at least once or always, respectively.
Moreover, the construction underlying our work also considers \emph{strong release} $\phi \ltlSrelease \psi \equiv \psi \ltlUntil (\psi \land \phi)$, \emph{(weak) release} $\phi \ltlRelease \psi \equiv \ltlGlobally \psi \lor (\phi \ltlSrelease \psi)$, and \emph{weak until} $\phi \ltlWuntil \psi \equiv \ltlGlobally \phi \lor (\phi \ltlUntil \psi)$.
Considering these additional operators allows formulas to be represented in \emph{negation normal form}, i.e.\ the negation $\lnot$ only appears in front of atomic propositions.
In the interest of space, we refer to \cite{DBLP:journals/jacm/EsparzaKS20} for precise definition on the semantics and discussion of these subtleties.
Understanding these issues is however not required for this work.

%Given an LTL formula $\phi$, the set of its \emph{sub-formulae} is denoted by $\subformulas(\phi)$.
%The \emph{top-level temporal operators} $\topformulas(\phi)$ are all temporal operators not nested inside other temporal operators.
%For example, the formula $\phi = \ltlGlobally ((\ltlFinally a) \land b) \land \ltlFinally b$ has sub-formulae $\subformulas(\phi) = \{a, b, \ltlFinally a, (\ltlFinally a) \land b, \ltlGlobally ((\ltlFinally a) \land b), \ltlFinally b, \phi\}$ and top-level operators $\topformulas(\phi) = \{\ltlGlobally ((\ltlFinally a) \land b), \ltlFinally b\}$.

\paragraph{LTL Synthesis} is an instance of the general synthesis problem, where the specification to be satisfied is given in form of an LTL formula \cite{DBLP:conf/icalp/PnueliR89}.
Due to recent advances \cite{DBLP:conf/atva/GaiserKE12,DBLP:conf/atva/KretinskyL13,DBLP:conf/atva/KomarkovaK14,DBLP:conf/cav/KretinskyMSZ18,DBLP:conf/lics/EsparzaKS18,DBLP:journals/jacm/EsparzaKS20}, the \emph{automata-based approach} \cite{DBLP:conf/lics/VardiW86} to LTL synthesis received significant attention.
In particular, the tool \texttt{Strix} \cite{DBLP:conf/cav/MeyerSL18}, built on top of \texttt{Owl} \cite{DBLP:conf/atva/KretinskyMS18}, which in turn implements these ideas, won several iterations of the synthesis competition SYNTCOMP \cite{DBLP:journals/corr/abs-2206-00251}.
Essentially, the given LTL formula is translated into an $\omega$-automaton, which in turn is transformed into a parity game.
Solving the resulting game yields a solution to the original synthesis question.

This game is obtained by \enquote{splitting} the automaton, as follows.
The set of atomic propositions is split into system- and environment-controlled propositions, i.e.\ $\AP = \AP_\sys \union \AP_\env$, and the players' actions correspond to choosing which of their propositions to enable.
Once both players chose their propositions' values, the automaton moves to the next vertex according to the players' choices.
Concretely, for an automaton state $p$, the environment can choose to move into $(p, v)$ where $v \subseteq 2^{\AP_\env}$, and from there, system can move to any automaton state $q = \delta(p, v' \union v)$ where $v' \subseteq 2^{\AP_\sys}$ and $\delta$ is the transition function of the automaton.
In particular, this means that the obtained game is \emph{alternating}, i.e.\ system and environment take turns in alternation.
Moreover, by convention the environment moves first.
See e.g.\ \cite{DBLP:conf/cav/MeyerSL18} for more details on this approach.

\paragraph{Semantic Translations} from LTL to automata are the key ingredient to our approach.
On top of providing a parity game, they also give a \emph{semantic labelling}, i.e.\ interpretable meaning, to the game's vertices.
In particular, the %approaches introduced in \cite{DBLP:conf/atva/KomarkovaK14,DBLP:conf/lics/EsparzaKS18} \cite{cav12}
approach introduced in \cite{DBLP:conf/cav/EsparzaK14} (see also \cite{DBLP:conf/lics/EsparzaKS18,DBLP:journals/jacm/EsparzaKS20,DBLP:journals/sttt/EsparzaKRS22})
and implemented in \texttt{Owl} \cite{DBLP:conf/cav/KretinskyMSZ18} intuitively yields for each vertex a list of LTL formulae, which roughly correspond to (sub-)goals which still have to be fulfilled, possibly repetitively.

\subsection{Our Goal}
\begin{figure}[t]
	\centering
	\begin{tikzpicture}[auto]
		\node[systemltl, initial above,align=center] (1) at (0,0) {$a \land \ltlNext \ltlGlobally b$ \\ $\lor$ \\ $\lnot a \land \ltlFinally c \land \ltlGlobally(r \rightarrow \ltlFinally g) \land \dots$};
		\node[anchor=east,systemltl] (2) at (-3.5,0) {$\ltlGlobally b$};
		\node[anchor=west,systemltl] (3) at (3.5,0) {$\ltlFinally c \land \ltlGlobally(r \rightarrow \ltlFinally g) \land \dots$};
		
		\path[->]
			(1) edge[swap] node {$a$} (2)
			(1) edge node {$\lnot a$} (3)
		;
	\end{tikzpicture}
	\caption{
		Motivational example to provide guidance through semantic labelling.
	}\label{fig:guidance}
\end{figure}
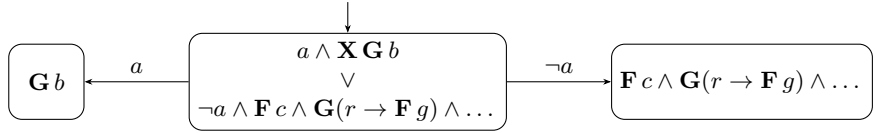

In this work, we want to demonstrate that this semantic labelling can be efficiently exploited for reactive synthesis.
For a motivational example to consider semantic labelling, we display a (vastly simplified) labelled game in \cref{fig:guidance}.
We are offered with the choice of choosing $a$ or $\lnot a$.
While it is not completely clear that choosing $a$ is indeed better, it certainly seems to be more promising, as the subsequent labelling seems much \enquote{easier} to handle.
Thus, faced with a choice, we likely would first try to win with $a$.
Observe that without the semantic labelling, our best option in this situation would be a random guess.
In \cite{DBLP:conf/atva/KretinskyMM19}, the authors used a simple, manually designed mechanism trying to capture this notion, called \emph{trueness}.
Motivated by the (surprisingly good) results of this approach, we want to tackle this problem by more sophisticated means.
Concretely, we want to make meaningful decisions based on the the labelling.
However, while the theory underpinning semantic translations is quite clean and pleasant \cite{DBLP:journals/jacm/EsparzaKS20}, the actual labellings appearing in practice are quite complex.
To further complicate things, the highly optimized implementation thereof \cite{DBLP:conf/cav/KretinskyMSZ18} employs several subtle optimizations and special cases.
We provide an example to showcase the complexity of this labelling in practice later in \cref{sec:handling}, kept brief in the interest of space, and a small real-world example in \iftoggle{arxiv}{\cref{sec:appendix:realistic_example}}{the Appendix of \cite{arxivVersion}}.
Since we have a simple intuition which however seems difficult to formalize, we opt to tackle this problem through means of machine learning.

\section{Previous Approaches and Their Limitations}

In this section, we briefly summarize the ideas of \cite{DBLP:conf/atva/KretinskyMM19} and the inherent problems associated with them.
The primary motivation of \cite{DBLP:conf/atva/KretinskyMM19} is to exploit the semantic labelling provided by \cite{DBLP:conf/cav/KretinskyMSZ18}, which gives us an indication of the long term goals in the game.
As an analogy, consider the game of chess.
Here, the \enquote{semantic labelling} is given by the board state, i.e.\ the position of each piece.
This labelling provides us with a reasonable indication of (i)~our current situation and (ii)~which moves might be better than others.
In particular, understanding and evaluating the semantics of the game is what allows humans to have a good intuition about the quality of moves, without thinking through the intractably large game tree.
Likewise, this understanding is what enabled algorithms to perform beyond human capabilities.

\subsection{Parity Game Solving by Trueness}
A central notion of \cite{DBLP:conf/atva/KretinskyMM19} is \emph{trueness}, an approximation of how close a formula is to being satisfied, i.e.\ $\true$.
The intuition is that the semantic labelling of states effectively describes \enquote{goals} of the system player.
If the formula is $\true$, the system has satisfied all goals and consequently won the game.
Likewise, increasing the trueness is indicative for a good move.
Remaining with the analogy of chess, trueness somewhat corresponds to counting the number of pieces on the board (or rather the difference between our and the opponent's pieces):
If no enemy pieces remain, we certainly have won, and a change of this difference, i.e.\ capturing an enemy piece or avoiding capture of own pieces, is a good indicator for the quality of a move.
In particular, this prohibits us from taking moves which immediately lead to a piece being taken.

In \cite{DBLP:conf/atva/KretinskyMM19}, the authors propose two ideas.
First, they suggest to use a trueness-maximizing strategy as initial one for strategy iteration, i.e.\ in each state select the edge which maximizes (or minimizes, in the case of $\env$) the obtained trueness.
Second, they use \emph{Q-Learning}, a popular reinforcement learning approach, as a solver for parity games, i.e.\ as competitor to strategy iteration, using three different reward signals.
There, each edge is given a reward, which is mostly based on (the change of) trueness, and these values then are back-propagated until choosing optimal rewards in each step yields a winning strategy.

While they also show Q-Learning to be an interesting avenue, we primarily focus on the \enquote{initializing strategy iteration} approach, since our goal is to augment exiting strategy iteration solvers.
Moreover, the experimental evaluation of \cite{DBLP:conf/atva/KretinskyMM19} suggests that Q-Learning scales poorly to large real-world formulae.

\subsection{Problems}
We now outline two key issues of this approach. %, which partly already have been mentioned in \cite{DBLP:conf/atva/KretinskyMM19}.
\begin{description}
	\item[Myopic Trueness]
	The primary heuristic in \cite{DBLP:conf/atva/KretinskyMM19} is trueness.
	While this approach already performs surprisingly well, especially for so called \emph{safety} and \emph{co-safety} formulae, it fails to take into account temporal dependencies; trueness is myopic.
	Again, considering chess, while counting the change of pieces does help us avoid \enquote{obviously stupid} moves, it does not stop us from moving pieces into positions where they are effectively guaranteed to be taken eventually and does not allow for sacrificing a piece in exchange for a long-term advantage.

%	\item[Dependence on Normalization]
%	%
%	As another signal, they consider the assigned priorities (however only for Q-Learning).
%	The reward signal derived from priorities requires normalization into the interval $[-1, 1]$, which in turn requires knowing the largest occurring priority.
%	For the concrete construction used by \cite{DBLP:conf/atva/KretinskyMM19}, this means that the complete game needs to be built, which quite often is the main bottleneck in LTL synthesis.
%	In particular, this rules out using their idea of priority-based feedback for on-the-fly approaches. \todo{We can remove this if we need space}

	\item[Manual Design]
	Their reward functions were defined manually, in contrast to being obtained from a learning process.
	While the intuition behind these definitions is reasonable, obtaining a guidance heuristic as a result of an optimization process is a much more principled approach.
\end{description}
We proceed to outline how we tackle these issues by a more sophisticated approach.

\section{A New Hope} \label{sec:ground_truth}

We want to improve reactive synthesis by applying machine learning.
As already motivated by \cite{DBLP:conf/atva/KretinskyMM19}, we want to approach this problem by identifying \enquote{promising} edges, choosing those as initial strategy for SI.
Naturally, as a first step, we need training data for our learning approach.
In particular, we need to identify which actually are the actual good choices in games, i.e.\ the \emph{ground truth}.
As it turns out, this is more complicated than one might expect.

\subsection{Obtaining Training Data with SI}
As SI allows us to solve a game and determine winning edges, one might try to employ SI for obtaining a ground truth (as we did initially).
However, SI actually provides us with potentially misleading or even conflicting information!
As we already hinted in the introduction through \cref{fig:SI_bad}, SI cannot give us a canonical ground truth.
In the example, one edge is winning iff the other is not used, and vice versa.
Thus, SI will yield a strategy which does not take both edges and we would consider one of them losing.
Moreover, note that there is no fundamental reason to prefer one edge over the other, so SI might in one run classify the edge from $v_2$ to $v_3$ as good and in a second run (or on a similar game) do the opposite or even consider neither winning.
%
%Here, either $v_2$ or $v_3$ may play to the respective other state, however we cannot do both, as then the target $v_\top$ will not be reached and the system loses.
%
%%Note that declaring both edges as losing is not good either, as this makes them indistinguishable from the self loops of $v_2$ and $v_3$, which are certainly worse.
%
%%While not optimal, it is certainly better to try winning via the respective other state than looping and losing for sure.
The underlying problem is that parity games do not allow for a unique \emph{maximally permissive} strategy (see e.g.\ \cite{DBLP:journals/ita/BernetJW02}), thus we cannot derive the \enquote{suitability} of an edge from a single solution strategy.

%Already for simple reachability formulae like $\ltlFinally (a \mathop{\mathsf{xor}} \ltlNext b)$, where both variables are controllable by the system, problems arise.
%The corresponding game is depicted in \cref{fig:SI_bad}.

\subsection{Solving the Game Tree}
Instead of using a particular strategy obtained from SI, we therefore propose to identify \enquote{all} solutions, i.e.\ all edges which are part of a winning strategy.
More formally, for each vertex $v$ we want to determine the value of each outgoing edge in the corresponding game tree rooted at $v$.
To prefer \enquote{shorter} solutions over larger, we add a beta-decay to the value.
Concretely, suppose we consider the game tree state $s = (v_1, \dots, v_i)$ which ends in a system state $v_i$.
Then, the value of $s$ is defined by $\mathrm{val}(s) = \beta \cdot \max_{s' \in \text{successors}(s)} \mathrm{val}(s')$ for a fixed $0 < \beta < 1$.

As we already mentioned, evaluating this tree is intractably large, namely exponential in the size of the game, which itself is already doubly-exponential in the input formula \cite{DBLP:conf/mochart/KupfermanR10,DBLP:conf/icalp/PnueliR89}.
Thus, we employ a classical technique of game theory.

\subsection{Monte Carlo Tree Search (MCTS)} \label{sec:truth:mcts}
Intuitively, we explicitly unfold the tree up to a specified depth, e.g.\ 7 plies, and then assign the results of (guided) random sampling to the occurring leaves, approximating the (beta-decayed) value of the game in these vertices.

We describe our method to approximate the value of a node $s = (v_1, \dots, v_i)$ in the game tree.
In essence, starting from $v_i$, we randomly select successors, with the following restrictions for each player.
The environment plays \emph{optimally}, i.e.\ if a state is winning for the environment (which we can determine beforehand through classical approaches) we immediately stop sampling and return a value of $0$.
Otherwise, the environment heuristically tries to delay the play as long as possible (decreasing the value the system player obtains due to beta-decay).
In contrast, the system player checks in a one-step lookahead if a choice is trivially winning, i.e.\ leading to a state labelled $\true$, always choosing such an edge if one exists.
Otherwise, the system randomly chooses among edges which are not trivially losing, i.e.\ lead to a $\false$ state.
If either player closes a loop, i.e.\ selects a successor which already occurs along the path, we determine the value by checking if the loop is winning or losing.
A loss yields a value of $0$, while a win yields $\beta^{\text{length}}$.
In summary, we approximate the probability of winning by playing randomly (avoiding obvious mistakes) against an optimal opponent, under-approximating the true value.
%Thus, for a large enough number of samples, the empirical expectation of this process converges to the true value with probability 1.
We deliberately opt for this random-choice approach to prefer regions where there is less potential for error.

\subsection{Optimizations}

While MCTS makes approximation of the game tree value feasible, we added several further technical improvements to arrive at a practically viable method.

\paragraph{SCC Decomposition}
We exploit the structure of the game by decomposing it into its strongly connected components (SCCs) and put them in reverse topological order.
Computing (or approximating) the value in that order allows for caching:
Once a run in the game tree leaves an SCC, it can only reach SCCs further down in the topological order, and, since we compute values in this order, the value of the reached state is already known, allowing us to re-use it immediately.
%, which speeds up the overall process and even implicitly increases the number of simulations.
%In particular, once a simulated run leaves the SCC it originated in, it reaches a state that has already been completed, due to the ordering.
%Using that states Monte Carlo value as result of the simulated run, means that this run can terminate and that it incorporates the information of many previous runs.

\paragraph{Pruning}
In addition to employing the MCTS values as game values in the tree expansion, we also use it to prune the game tree.
In particular, once we computed the Monte Carlo values for each state, we restrict the choice of the environment to the successors which yield (close to) the lowest Monte Carlo value (recall that the environment prefers lower values).
We empirically chose $0.02$ as a threshold, i.e.\ we only keep those edges for the environment which are within $0.02$ value of the lowest decision.
While in theory this might remove crucial paths due to statistical fluctuations of MCTS, in practice it allows for a much deeper game tree, which in our experiments heavily outweighed the theoretical downside.
\section{Handling the Truth} \label{sec:handling}

We introduced a way how to obtain a well-founded notion of \enquote{value} (to be precise, an approximation thereof) for a choice, i.e.\ an indication how good this choice is.
%Now, we can determine for each game an (approximation of) the value in each edge, i.e.\ how 
As such, we can rank edges by their value in each state.
Intuitively, picking an edge which is ranked very highly should correspond to a good chance of winning.
A high value means that even against an optimal player we can very likely close a winning loop, and, due to beta decay, do so quickly, thus minimizing the chance for an error.
%As a side effect, this also implies that for edges with high value there is less chance of error, since the winning loop can be closed quickly.

Recall that our goal is to provide a good initial strategy.
Thus, the exact values actually are irrelevant, since we only want to give the best edge as initial choice.
Instead of trying to predict the exact value, we therefore want to learn this relative ranking.
Formally, suppose we consider a system vertex $v \in \VerticesSys$ with edges $E_v = \{(v, u) \mid (v, u) \in E\}$.
A ranking of edges effectively corresponds to a (total) order ${\prec_v} \subseteq E_v \times E_v$.
The principle of \emph{pairwise ranking} \cite{liu_09_LTR} suggests that we learn a function $f : E_v \times E_v \to \{-1,1\}$ that classifies pairs of edges depending on which one is the better choice, i.e.\ $f(e, e') = 1$ if $e' \prec_v e$ and $-1$ otherwise.
However, such a function might not be perfect.
For example, we could get $f(e_1, e_2) = 1$, $f(e_2, e_3) = 1$, and $f(e_3, e_1) = 1$, which is incompatible with any order.
Thus, learning to rank suggests to determine an ordering $\prec$ that minimizes the \emph{inversions} w.r.t.\ $f$, i.e.\ the number of cases where $f(e, e') = 1$ but $e \prec_v e'$.
This problem, called rank aggregation, is known to be $\NP$-hard, and we employ a greedy approximation as suggested by \cite{liu_09_LTR}.

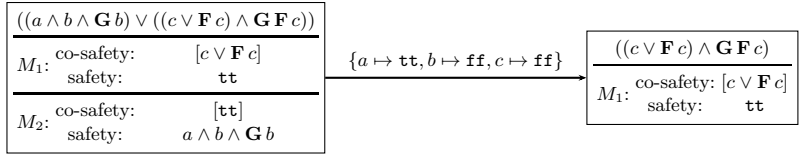
\begin{figure}[t]
	\centering
	\begin{tikzpicture}[every node/.style={scale=0.8}]
		\node[rectangle,draw] (q0) at (0,0) {
			\begin{tabular}{lcc}
				\multicolumn{3}{c}{ $((a \land b \land \ltlGlobally b) \lor ((c \lor \ltlFinally c) \land \ltlGlobally \ltlFinally c))$} \\
				\midrule
				\multirow{2}{*}{$M_1$:} & co-safety: &                             $[c \lor \ltlFinally c]$                              \\
				                        &  safety:   &                                      $\true$                                      \\
				\midrule
				\multirow{2}{*}{$M_2$:} & co-safety: &                                     $[\true]$                                     \\
				                        &  safety:   &                         $a \land b \land \ltlGlobally b$
			\end{tabular}
		 };
		\node[rectangle,draw] (q1) at (7,0) {
			\begin{tabular}{lcc}
				\multicolumn{3}{c}{$((c \lor \ltlFinally c) \land \ltlGlobally \ltlFinally c)$} \\
				\midrule
				\multirow{2}{*}{$M_1$:} & co-safety: &         $[c \lor \ltlFinally c]$          \\
				                        &  safety:   &                  $\true$
			\end{tabular}
		};
		
		\path[thick,->] (q0) edge node[above] {$\{a \mapsto \true, b \mapsto \false, c \mapsto \false\}$} (q1);
	\end{tikzpicture}
	\caption{
		A single transition in the automaton computed for the formula $(a \land \ltlGlobally b) \lor \ltlGlobally \ltlFinally c$.
	} \label{fig:example_labelling_complicated}
\end{figure}

Our concrete goal thus now is to learn such a function $f$ based on the semantic labelling of the start and end vertices of the two edges.
We want to employ machine learning for this purpose:
While the high-level intuition of the semantic labelling is rather clear, the actual implementation used to obtain the games \cite{DBLP:conf/atva/KretinskyMS18} employs numerous optimizations, separate cases, etc.
To provide the reader with a sense of the complexity, we display a single edge in the automaton obtained for a simple formula in \cref{fig:example_labelling_complicated}, and a real-world scenario in \iftoggle{arxiv}{\cref{sec:appendix:realistic_example}}{ the Appendix A.1 of \cite{arxivVersion}}.

We proceed to describe (i)~(some of) the features we use, i.e.\ which quantities we extract from the labelling, (ii)~the model we employ, and (iii)~the dataset and methodology used to train our model.
%In particular, we mention (some of) the features we provide to our learning approach, the model we use for learning, and finally a description of the training process.

\subsection{Features} \label{sec:handling:features}
In total, we have defined over 200 different features to convert the edges into a usable vector of reals.
In the interest of space we only present the high-level ideas of a small subset which covers most interesting ideas.
%Recall that none of the features tries to be a good heuristic on its own and there will always be examples where a feature might be misleading.
%It is only the correct combination of multiple features that results in a good heuristic.

Since most information is contained in the states rather than in the edges themselves, the majority of our features are defined for the former.
An edge is then either associated with the feature value of its successor or with the change in a feature value between its predecessor and successor.
As indicated in \cref{fig:example_labelling_complicated}, the semantic labelling comprises several formulae, namely a \enquote{master} formula, which intuitively indicates the global state, and several \enquote{monitors} (which themselves comprise several formulae), monitoring repeating sub-goals.
We define \emph{base features}, which convert a single formula to a single number.
These features can then be applied to both the master as well as monitor formulae, where further aggregation is necessary.
%As the order of the monitors resembles their significance, we employ for weighted average.
Some notable base features are the following:
\begin{description}
	\item[Number of Conjuncts]
	We count the number of conjuncts if the top level operator is a conjunction and otherwise default to 1.
	The intuition behind this feature is that less conjuncts tend to correspond to a less constrained formula.
	Further, reducing the number of conjuncts along an edge often means that sub-goals have been achieved.
	(We consider several further syntactic features such as the number of disjuncts, the height of the syntax tree, or the number of temporal operators, which all follow similar ideas.)

	\item[Trueness]
	Since this has proven to be a solid heuristic on its own, we again incorporate it as a feature.
	
	\item[System Control]
	This feature (and variations thereof) incorporate the information of the variable partitioning by approximating how much impact the choice of the system variables has on the truth value of the formula.
	Intuitively, a higher system control is desirable.
	Further, this feature also counteracts false positives of, e.g., trueness, as high values of trueness are worth much less if the system has no control on whether one of the many satisfying assignments is played.

	\item[Obligation Set]
	This group of features is based on the idea of \emph{obligation sets} as introduced by \cite{10.1109/TIME.2013.19}.
	In essence, an obligation set for a formula $\varphi$ is an assignment that, if played indefinitely, satisfies the formula. %, i.e.\ $\someAssignment^\omega \models \varphi$.
	Using the inductive definition of \cite{10.1109/TIME.2013.19}, we can compute a formula $\varphi'$ whose satisfying assignments are exactly the obligation sets of $\varphi$, see \iftoggle{arxiv}{\cref{app:osf_def}}{Appendix A.2 of \cite{arxivVersion}}.
	Using this new formula, we can obtain numerous new features by applying other base features to $\varphi'$.
	In particular, we are interested in the new formulas trueness as this indicates how many obligation sets exist.
	Further, we are interested in its system control, as a higher value makes it more likely that the system can enforce at least one obligation set.
\end{description}
In addition to the base features, we define the following edge-specific features:
\begin{description}
	\item[Priority]
	As priorities are crucial for winning a play, it is only natural to incorporate that information in our features.
	However, as SVMs struggle with parity information, we reorder the priorities by how beneficial they are for the system and map them to $[-1,1]$ (similar to \cite{DBLP:conf/atva/KretinskyMM19}).
	In particular, the smallest odd priority gets mapped to $1$ and the smallest even priority to $-1$.
	For this normalization, we use an a-priori upper bound provided by the underlying automaton construction.
%	This bound was not available for \cite{DBLP:conf/atva/KretinskyMM19}.

	\item[Progress]
	This feature is rather similar to \cite{DBLP:conf/atva/KretinskyMM19}'s progress feature.
	We compute the percentage of already succeeded sub-goals of a monitor (instead of their trueness) and aggregate by weighted average (rather than maximum).
	Additionally, we introduce punishments for failing monitors. % and slightly smaller punishments for failing monitors that reappear ($\ltlGlobally\ltlFinally$-monitors).
	Intuitively, this encourages long-term progress for temporal goals.

	\item[One Step]
	Here, the idea is to recommend an assignment that is to be played in the current state by traversing the syntax tree and propagating recommendations upwards, which is inspired by message passing in graph neural networks.
	For example, if we see $a \land b$ we strongly recommend playing $a$ and $b$, if we see $\ltlFinally (a \land b)$ we take the previous recommendation and tune it down, since $\ltlFinally$ is \enquote{less urgent}.
	The feature value is obtained by measuring how well the valuation of an edge aligns with the recommended assignment.
%	Note that the recommendation cannot be particularly good by itself, as otherwise it would single-handedly solve the overall problem of recommending good edges.
%	However, how well an edge aligns with this possibly bad recommendation, yields new orthogonal information in which the SVM can find correlations in.
\end{description}

\subsection{Pair Classification by Support Vector Machines} \label{sec:handling:svm}

To instantiate our pair classification function $f$, we opt for support vector machines.
In principle, one could employ any binary classifier, which is why we also experimented with other models such as decision trees, random forests or gradient boosted trees.
However, SVMs proved to perform best, which we attribute to their great ability to generalize due to their margin maximizing nature \cite{liu_09_LTR}.
Additionally, SVMs are rather simple (compared to our other options) and provide us with extra information known as \emph{confidence}.
Given by the distance of the predicted sample to the decision hyperplane, its magnitude can be interpreted as how confident the SVM is in its prediction.
We denote the confidence of a pair $(e_1,e_2)$ by $c(e_1, e_2)$ and use it to slightly alter the greedy ranking algorithm from literature.
To rank the edges of a vertex $v$, each edge $e \in \Edges_v$ gets assigned a score $s(e) = \sum_{e' \in \Edges_v, e'\neq e} c(e,e')$.
Recall that if we predict $e \prec_v e'$, the confidence is negative.
Finally, we rank the edges according to their score, where a higher score corresponds to a better edge, and the recommended strategy is obtained by playing the highest ranked edge for each state.

\subsection{Further Notes on Implementation}
In addition to the feature extraction, there are several other engineering aspects, which are crucial for the final performance.
In this section, we comment on the three most important ones.

\paragraph{Statewise Feature Normalization}
Before passing the features to the model, we proceed to normalize them.
Due to possible future applications in on-the-fly solvers, we only consider feature values of edges from the same state for this normalization.
The crucial observation is that this already introduces comparative information in the features.
A normalized trueness value of 1, for example, means this edge has the best trueness among all other edges from their state although it does not tell us anything about its absolute value.
While the latter might also be important in theory, we observed that in practice the statewise normalized value is more important with only a few exceptions.

\paragraph{State Classification}
We observed several significantly different behaviours required in different states.
For example, in some states we need to exclusively focus on the master formula, while in others only the monitors play a role.
This also relates to the underlying principles of the automaton construction.
%Not all game states are created equally!
%There are some whose purpose is advancing the master formula, while others should only focus on their monitors and closing accepting cycles.
It is very difficult, especially for a simple model like an SVM, to switch between different behaviours.
We divide states into three groups which approximate the different classes, and train separate models for each class.
The three classes we suggest are (i)~states without monitors, (ii)~states where the master formula does not change in any successor, (iii)~and states that fall into neither category.
In addition to having the separate models learn separate behaviours, we can also provide them with separate feature sets that only include relevant information.
For example, the first class only requires features of the master formula, whereas these can be neglected in the second one.

\paragraph{Complement Construction}
The underlying automaton construction uses the fact that the system being able to enforce satisfaction of a formula $\varphi$ is equivalent to the environment being able to enforce falsification of $\lnot \varphi$.
In other words, solving the game for the negated formula and swapped roles yields the same result.
However, in the game obtained for $\lnot \varphi$ the role of \enquote{system}, the player who choses second and for which we learnt the recommendation, i.e.\ for transitions from states $(p, v)$ to $q$, now corresponds to the original environment.
This drastically changes the meaning of features.
For example, a trueness of 0 suddenly is very desirable.
We tackle this by training separate models for both cases.
Together with state classification, this yields a total of 6 different models that we assemble for our heuristic.

\subsection{Training the Model} \label{sec:handling:training}
With these ideas at hand, we conclude this section by discussing our dataset, in particular how we preprocess it, and how we train our model.

\paragraph{Dataset and Preprocessing}
As one of our goals is to exploit human bias in writing LTL formulae, the foundation of our dataset is given by the LTL benchmarks of SYNTCOMP. \footnote{Available on GitHub \url{https://github.com/SYNTCOMP/benchmarks}.}.
To further augment the data, we mutate these formulae by randomly replacing temporal operators.
This yields new (random) samples that syntactically resemble the original, human-written structure.
For practical reasons, we only consider formulae which can be converted to a DPA within 10 minutes. %using the LDBA method \cite{Sickert_16_LDBA,DBLP:conf/tacas/EsparzaKRS17} 
Ultimately, this leaves us with 405 original and 514 mutated formulae, of which we use 60\% each for training, 20\% for validation, and 20\% for evaluation, see \cref{sec:experiments}.

Obtaining the edge pairs for training requires several further steps.
First of all, we exclude trivial cases that can easily be detected by simple rules (see \cref{sec:truth:mcts}), allowing our model to focus on complicated cases. %, i.e.\ pairs whose predecessor is trivially won or where one of its edges is trivially lost.
%As these cases can be easily detected and dealt with manually upfront, the model can put more focus on the relevant cases.
Further, we exclude pairs where the ground truth value happens to be equal, as it is unclear which edge the model should predict.
In particular, we exclude all edges originating in losing states (since there is no sensible action to recommend).
Finally, we only include a limited amount of pairs per game in the training set:
Pairs of the same game tend to look similar, thus a few disproportionately large games would result in a very unbalanced dataset.
All remaining edge pairs are added in both orders, i.e.\ $((e_1,e_2),y)$ and $((e_2,e_1),-y)$, where $y\in\{1,-1\}$ determines which edge is better, in order to prioritize teaching symmetry to the model.

\paragraph{Training}
For each of the 6 models, we first compute mean and standard deviation of the respective training set and use them to standardize the input to $\mathcal{N}(0,1)$.
Further, we perform recursive feature elimination for each state class individually, adapted to features appearing twice (once for each input edge).
For each state class, we ended up with 30-40 features.
%We note a subtle twist:
%Each feature actually appears twice in the input, as we consider pairs of edges.
%So, if we want to eliminate a feature, we would remove it for both edges from the input.

For the actual training process, we performed an extensive grid search for several model types (decision trees, random forests, etc., see \cref{sec:handling:svm}) in order to determine suitable values for the hyper-parameters.
As mentioned earlier, we ultimately opted for the SVMs due to their simplicity and generalization abilities.

%\todo{Interesting Fact: DT of depth 2 is enough to correctly classify 90\% of training pairs in class 3(now class 2) states?}

%\input{4_implementation}
\section{Experimental Evaluation}\label{sec:experiments}

In this section, we present experimental evaluation of our tool \texttt{SemML}.
The model was learnt by communicating the relevant data to a Python process running \texttt{scikit-learn} \cite{scikit-learn}.
We then extracted the learnt weights and, based on them, implemented the recommendation procedure in Java, on top of \texttt{Owl} \cite{DBLP:conf/atva/KretinskyMS18}.
The artifact can be found at \cite{artifact_newest}, which references a slightly improved version from the one we submitted to the artifact evaluation \cite{artifact_eval}.

\subsection{Evaluation Goals}
Our primary goal in this work is to show that our approach, enabled by our new ground truth, can be used to solve more complicated instances than the approach of \cite{DBLP:conf/atva/KretinskyMM19}, in particular formulae going beyond pure (co-)safety.
Thus, our first evaluation goal is the following:
\begin{quote}
	\emph{Research Question 1:} How much does our model based on SVM and the game tree ground truth outperform the trueness-based initial strategy recommendation approach of \cite{DBLP:conf/atva/KretinskyMM19}?
\end{quote}
We refer to the trueness-based initial strategy of \cite{DBLP:conf/atva/KretinskyMM19} as \texttt{TrueSI}.

Although not the focus of this work, we ultimately want to improve synthesis through meaningful exploration guidance, in particular, by suggesting likely winning edges.
Thus, we are interested how our prototype performs in a real-world scenario.
\begin{quote}
	\emph{Research Question 2:} How do initial strategies recommended by our approach synergize with state-of-the-art synthesis tools?
\end{quote}
We address both questions separately.

\subsection{RQ1: Quality of Initial Strategy} \label{sec:experiments:atva}
\paragraph{Datasets}
To fairly compare to \cite{DBLP:conf/atva/KretinskyMM19}, we consider the same dataset, i.e.\ randomly generated LTL formulae, split into three categories: \enquote{(Co-)Safety}, \enquote{Near (Co\nobreakdash-)Safety}, and \enquote{Parity}.
%The first contains, as the name suggests, (co\nobreakdash-)safety formulae.
%The second class are formulae \enquote{close to} being (co-)safety and only have small variations in them.
%Finally, the third class are arbitrary formulae, without any restriction.
See \cite{DBLP:conf/atva/KretinskyMM19} for details on how these are obtained.
In essence, the tool \texttt{randltl} \cite{DBLP:conf/atva/Duret-LutzLFMRX16} is used to generate random formulae with different biases.
Then, we filter out formulae which need more than 10 minutes to be translated to a parity automaton.
As a second dataset, we also use some (original and mutated) SYNTCOMP formulae (the test set described in \cref{sec:handling:training}).
We only consider formulae where the corresponding game can be won by system.
We do this simply because we can only recommend on games which are winning -- otherwise there is no preference on edges since every action is losing by definition.
In total, this leaves 262 randomly generated formulae and 123 from SYNTCOMP. %(58 original and 65 mutated ones).
%We deliberately do not put too much emphasis on randomly generated formulae, since our goal is to exploit human biases in designing LTL formulae.

\paragraph{Metrics}
We consider two metrics for our comparison.
Firstly, similar to \cite{DBLP:conf/atva/KretinskyMM19}, we consider the fraction of \emph{immediately solved} games, i.e.\ games where following actions recommended by \texttt{SemML} or \texttt{TrueSI} directly yields a winning strategy.
In light of our motivation to augment SI solvers, we want to measure how \enquote{close} the recommended strategy is to being correct in case is not immediately winning.
To this end, we feed it to (a modified version of) the parity game solver \texttt{Oink} \cite{DBLP:conf/tacas/Dijk18} and compute the \emph{(relative) distance} of the obtained strategy, as follows.
We count the number of (reachable) states in which the winning strategy determined by \texttt{Oink} differs from the recommended one, i.e.\ how many \enquote{wrong} choices were recommended, and divide it by the total amount of (reachable) states.
%The relative size refers to the portion of the game which is reachable under the obtained winning strategy.
%Intuitively, if this value is smaller then the winning strategy probably is (i)~simpler and (ii)~we need to construct a smaller part of the game in order to show that it is winning (which often is the bottleneck of synthesis).
We note that this unfortunately induces a slight bias that we cannot measure:
\texttt{Oink} may potentially change winning decisions because of internal details of the algorithm.
Ideally, we would want to obtain the minimal distance over all winning strategies; however this quantity is intractable to compute due to the exponential size of the strategy space.
Nevertheless, we believe that this measure strongly correlates with the quality of the strategy.

We argue that simply measuring the number of iterations required by strategy iteration to converge is a too crude metric:
On the one hand, even a \enquote{very wrong} strategy can be changed to a winning strategy in a single iteration by changing the choice in every single state.
On the other hand, even a nearly correct strategy, requiring only a hand full of changes, may need as many iterations.
Moreover, this additionally induces the same bias as above.
%Maybe the recommended strategy is close to being correct, but less \enquote{suitable} for \texttt{Oink}.

\paragraph{Expectations}
Since our approach incorporates trueness as one of its many features, we expect that our approach should be at least on par with the previous one of \cite{DBLP:conf/atva/KretinskyMM19}.
As we also consider long-term temporal information beyond trueness, we particularly expect to outperform \texttt{TrueSI} on larger, more complicated instances.

\paragraph{Results}
We ran this evaluation on consumer hardware (Intel Core i7-8565U with 16GB RAM).
We summarize our findings in \cref{tbl:rq1_summary}.
Clearly, our approach vastly outperforms the previous one.
In particular, while \texttt{TrueSI} perfectly handles (co\nobreakdash-)safety formulae, its performance quickly drops when going to more complicated formulae.
In comparison, the \texttt{SemML} solves the vast majority of formulae immediately, even on the quite complicated SYNTCOMP dataset.
We note that these findings are not \enquote{absolute} (as to be expected from machine learning approaches).
There are few instances where the previous approach does perform better.
Our baseline comparison to a random initialization approach %, which in all classes solves less than 10\% immediately, 
validates that both approaches indeed solve a non-trivial problem.

\begin{table}[t]
	\caption{
		Summary of our comparison between \texttt{TrueSI}, the approach of \cite{DBLP:conf/atva/KretinskyMM19}, and our tool \texttt{SemML}.
		We first list the fraction of immediately winning strategies (larger is better), followed by the geometric mean of the relative distance, i.e.\ the fraction of states in which the decision was adapted by \texttt{Oink} to obtain a winning strategy (smaller is better).
		For the first comparison, we also consider random initialization as a baseline.
		For this second comparison to be fair, we only consider games where neither tool yielded an immediately winning strategy.
	} \label{tbl:rq1_summary}
	\centering
	\setlength{\tabcolsep}{5pt}
	\begin{tabular}{ccccc}
		     Tool       &   (Co-)Safety   & Near (Co-)Safety &     Parity     &    SYNTCOMP    \\
		\midrule
		                       \multicolumn{5}{c}{Immediately Solving}                         \\
		\midrule
		\texttt{TrueSI} &      100\%      &       85\%       &      66\%      &      44\%      \\
		\texttt{SemML}  & \phantom{0}99\% &       95\%       &      88\%      &      85\%      \\
		    Random      & \phantom{00}7\% &  \phantom{0}2\%  & \phantom{0}5\% & \phantom{0}3\% \\
		\midrule
		                        \multicolumn{5}{c}{Relative Distance}                          \\
		\midrule
		\texttt{TrueSI} &       --        &       75\%       &      45\%      &      29\%      \\
		\texttt{SemML}  &       --        &       52\%       &      28\%      &      16\%      \\
		 Ratio of both  &       --        &       1.4        &      1.6       &      1.8
	\end{tabular}
\end{table}

\begin{figure}[t]
	\centering
	\begin{tikzpicture}[auto]
	\begin{axis}[
			width=0.5\textwidth,height=3.7cm,
			table/col sep=comma,
			xlabel=Size,ylabel=Solved \%,
			ymin=0,ymax=1,
			xmode=log,
			legend style={at={(0.5,1.1)},anchor=south}
		]
		\addplot+[draw=black,mark=x,thick,draw=gray,mark options={draw=black}] table [x index=0,y index=1] {data/RQ1/solvedPercentage_per_bin_woParity_svm.csv};
		%\addlegendentry{\texttt{SemML}}
		\addplot+[draw=black,mark=o,thick,draw=gray,mark options={draw=black}] table [x index=0,y index=1] {data/RQ1/solvedPercentage_per_bin_woParity_tru.csv};
		%\addlegendentry{\texttt{TrueSI}}
	\end{axis}
	\end{tikzpicture} %
%	\begin{tikzpicture}[auto]
%	\begin{axis}[
%			width=0.3\textwidth,height=4cm,
%			table/col sep=comma,
%			xlabel=Size,ylabel=Relative Size \%,
%			xmode=log,
%			legend style={at={(0.5,1.1)},anchor=south}
%		]
%		\addplot+[draw=black,mark=x,thick,draw=gray,mark options={draw=black}] table [x index=0,y index=1] {data/RQ1/relSize_per_bin_woParity_svm.csv};
%		%\addlegendentry{\texttt{SemML}}
%		\addplot+[draw=black,mark=o,thick,draw=gray,mark options={draw=black}] table [x index=0,y index=1] {data/RQ1/relSize_per_bin_woParity_tru.csv};
%		%\addlegendentry{\texttt{TrueSI}}
%	\end{axis}
%	\end{tikzpicture} %
	\begin{tikzpicture}[auto]
	\begin{axis}[
			width=0.3\textwidth,height=0.3\textwidth,
			table/col sep=comma,
			xlabel=\texttt{SemML},ylabel=\texttt{TrueSI},
			xmin=0,xmax=1,ymin=0,ymax=1
		]
		\addplot[black,forget plot,update limits=false] coordinates {(0,0) (1,1)};
		\addplot+[only marks,draw=black,mark=x,thick]  table [x index=2,y index=3] {data/RQ1/relDist_scatter_O_woParity.csv};
	\end{axis}
	\end{tikzpicture} %
	\caption{
		A detailed comparison on SYNTCOMP formulae.
		The left plot compares how many games were immediately solved, grouped by size and considering the (arithmetic) mean in each group.
		\texttt{SemML}'s values are displayed by crosses, \texttt{TrueSI} by circles.
		The right plot compares the relative distance of \texttt{SemML}'s and \texttt{TrueSI}'s solutions.
	} \label{fig:syntcomp_trueness_sizes}
\end{figure}
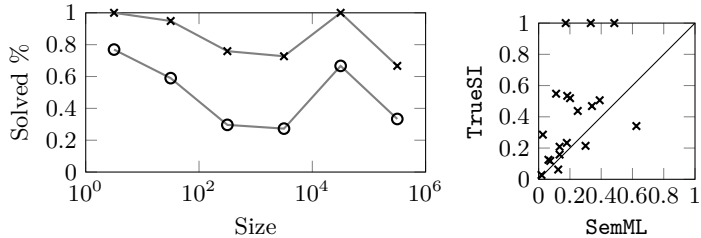

Since we are particularly interested in complex, \enquote{human written} formulae, we investigate the SYNTCOMP dataset more closely.
In \cref{fig:syntcomp_trueness_sizes}, we provide a more detailed view on our two metrics.
First, we investigate how the \enquote{immediately solving} performance evolves in comparison to the size of the game, which intuitively correlates with the difficulty of the synthesis question.
We observe that \texttt{SemML} solves practically all smaller games and still performs well on larger games, compared to \texttt{TrueSI}, which quickly falls off.
The second plot displays the relative distances for each instance which neither recommendation solved immediately.
We clearly see that the strategies recommended by \texttt{SemML} are better in almost all cases.

This positively answers our first question.
Aside from the direct comparison to the previous approach, the significant percentage of immediately solved games gives us an interesting implication:
If \texttt{SemML} solves many games immediately, we can use \texttt{SemML} as a best-effort guidance tool for reactive synthesis questions which are intractably large to solve.
Moreover, \texttt{SemML} thus presents us with a constant size representation of a winning strategy for many games, effectively described by approximately a few hundred SVM weights compared to a decision table for thousands of states in \emph{each} game.

\subsection{RQ2: On-the-fly SemML} \label{sec:experiments:strix}
In our second experiment, we evaluate the suitability of \texttt{SemML} for real-world parity game solving by using it as guidance tool for the state-of-the-art reactive synthesis tool \texttt{Strix} \cite{DBLP:conf/cav/MeyerSL18}.

\paragraph{Strix' Anatomy}
We first briefly describe how \texttt{Strix} works and how it uses guidance heuristics.
In essence, \texttt{Strix} builds the parity game on-the-fly, i.e.\ iteratively constructs parts of the game it deems important.
Then, two strategy improvements are running in parallel, one for either player.
Not yet explored states are treated as losing for both.
In this way, if we find a winning strategy for either player on the constructed part of the game, it is winning for the complete game.
Otherwise, we need to explore further.
%then by following that strategy this player guarantees that (i)~the sub-game is won by this strategy and (ii)~no unexplored state can be visited (otherwise, the strategy would not be winning).
%Thus, this is a winning strategy for the complete game.
Here, a key ingredient for practical efficiency is a heuristic to decide which states should be explored first:
If we explore states reachable under the \enquote{smallest} winning strategy, we naturally find this strategy as quickly as possible.
In its current form, \texttt{Strix} employs trueness for this guidance and selects an \emph{automaton} edge with the \emph{globally} highest trueness for exploration.
(Dually, edges with the lowest trueness are also followed, since these are \enquote{promising} for the environment.)

\paragraph{Integration}
We integrate \texttt{SemML} with \texttt{Strix} as follows.
Suppose we are asked to compute a global score for an automaton edge $e = (p, q)$ (recall that \texttt{SemML} gives \emph{local} advice on edges in the \emph{game}).
We explicitly build up the game between the automaton states $p$ and $q$, i.e.\ all choices of the environment in $p$ followed by the respective system choices.
For each occurring system state $s$, we compute the \texttt{SemML} ranking score as explained in \cref{sec:handling:svm}, i.e.\ the confidence based score.
This only gives us local information: the magnitude of our score only reflects the preference relative to actions available in the system state $s = (p, v)$.
Since the previously used trueness proved to be a good indicator for global progress, we multiply our local score by this global value.
Finally, to obtain a value for the automaton edge, we take the minimal value of all arising system states, since the environment chooses first.
We additionally apply straightforward rules such as assigning values of $0$ and $1$ values to $\false$ and $\true$ states, respectively.
Finally, \texttt{Strix} by default employs a decomposition approach, which does not build a single DPA.
Therefore, \texttt{SemML} would not be applicable, and we disable it for the purpose of  evaluation.

\paragraph{Dataset}
We considered 188 randomly selected formulae of SYNTCOMP (which were not used in the training of the model), also including unrealizable ones.

\paragraph{Metrics}
We evaluate the total required time to solve the game and compare to \texttt{Strix} in its normal configuration.
Since we expect the unoptimized computation of \texttt{SemML}'s advice to take considerable time, we separately measure the required time and additionally perform a comparison with this time subtracted.
Since our scoring function is a straightforward SVM, we strongly believe that by tailoring the evaluation to \texttt{Strix}' requirements, it can be significantly sped up.
In particular, our advice computation re-constructs information which is computed during the exploration of the automaton but difficult to access without significant changes to both \texttt{Strix} and \texttt{Owl}.

\paragraph{Expectations}
We do not expect this approach to work to its full potential because \texttt{Strix} architecture does not exactly fit our approach (recall that our primary motivation was to compare to \cite{DBLP:conf/atva/KretinskyMM19}).
We discuss these differences and possible ways to address them later.
Moreover, as we construct the intermediate game states for every recommendation and evaluate the recommender SVM several times, we expect that significant time is spent computing the advice of \texttt{SemML}.

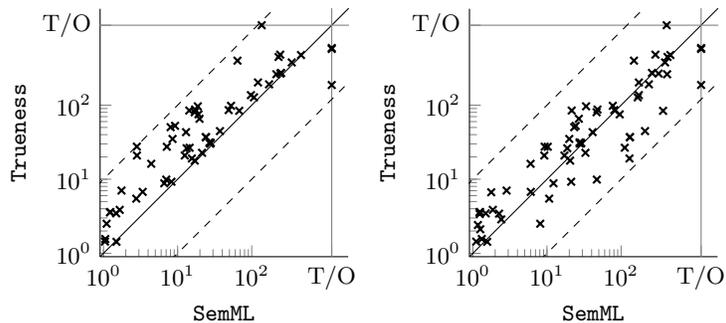
\begin{figure}[t]
	\centering
	\begin{tikzpicture}[auto]
	\begin{axis}[
			width=0.4\textwidth,height=0.4\textwidth,
			table/col sep=comma,
			xlabel=\texttt{SemML},ylabel=\texttt{Trueness},
			xtick={1,10,100},
			ytick={1,10,100},
			extra x ticks={1200},extra x tick labels={T/O},
			extra y ticks={1200},extra y tick labels={T/O},
			xmin=0.9,ymin=0.9,xmax=2000,ymax=2000,xmode=log,ymode=log,
			axis x line*=bottom,
			axis y line*=left
		]
		\addplot[black,forget plot,update limits=false] coordinates {(0.0000001,0.0000001) (1000000,1000000)};
		\addplot[gray,forget plot,update limits=false] coordinates {(0.0000001,1200) (10000,1200)};
		\addplot[gray,forget plot,update limits=false] coordinates {(1200,0.0000001) (1200,10000)};
		\addplot[black,dashed,forget plot,update limits=false] coordinates {(0.1,1) (100000,1000000)};
		\addplot[black,dashed,forget plot,update limits=false] coordinates {(1,0.1) (1000000,100000)};

		\addplot+[only marks,draw=black,mark=x,thick]  table [x index=1,y index=2] {data/RQ2/scatter_dpa_cleanTimeReal.csv};
	\end{axis}
	\end{tikzpicture} %
	\begin{tikzpicture}[auto]
	\begin{axis}[
			width=0.4\textwidth,height=0.4\textwidth,
			table/col sep=comma,
			xlabel=\texttt{SemML},ylabel=\texttt{Trueness},
			xtick={1,10,100},
			ytick={1,10,100},
			extra x ticks={1200},extra x tick labels={T/O},
			extra y ticks={1200},extra y tick labels={T/O},
			xmin=0.9,ymin=0.9,xmax=2000,ymax=2000,xmode=log,ymode=log,
			axis x line*=bottom,
			axis y line*=left
		]
		\addplot[black,forget plot,update limits=false] coordinates {(0.0000001,0.0000001) (1000000,1000000)};
		\addplot[gray,forget plot,update limits=false] coordinates {(0.0000001,1200) (10000,1200)};
		\addplot[gray,forget plot,update limits=false] coordinates {(1200,0.0000001) (1200,10000)};
		\addplot[black,dashed,forget plot,update limits=false] coordinates {(0.1,1) (100000,1000000)};
		\addplot[black,dashed,forget plot,update limits=false] coordinates {(1,0.1) (1000000,100000)};

		\addplot+[only marks,draw=black,mark=x,thick]  table [x index=1,y index=2] {data/RQ2/scatter_dpa_timeReal.csv};
	\end{axis}
	\end{tikzpicture}
	\caption{
		Scatter plot comparing \texttt{Strix} with guidance provided by \texttt{SemML} and the default \texttt{Trueness}.
		On the left, we depict the total runtime excluding time spent for computing the guidance, and on the right we show the total time.
		We plot all models for which at least one method produced a result and count timeouts as 20 minutes (twice the timeout of 10 minutes).
		Note that the plot is logarithmic.
		The dashed lines denote a 10x difference.
	} \label{fig:strix_timing}
\end{figure}

\paragraph{Results}
We conducted our experiments on a server with an Intel Xeon E5-2630 v4 processor with 256GiB of RAM and employed a 10 minute timeout per execution.
We summarize our findings in \cref{fig:strix_timing}.
Strikingly, our approach already performs favourably, despite the differences in architecture, hardly optimized advice computation, and no specific re-training for the task at hand.
Excluding the time spent for advice computation, our approach performs significantly better in practically all instances.
This answers our second question positively, too.

\subsubsection*{Adapting \texttt{SemML} to \texttt{Strix}}
In order to adapt our underlying approach, we require several non-trivial changes to \texttt{SemML}.
We discuss the \enquote{mismatches} between the current approach and how they could be addressed.
%However, there are several mismatches between our approach and the current architecture of \texttt{Strix}, which cannot be overcome easily.
First, \texttt{Strix} selects a globally optimal edge to explore while \texttt{SemML} suggest actions locally.
In particular, our scoring is not trained to compare edges of two different states.
While trueness seems to be a good compromise for the time being, we believe that (through significant engineering effort) \texttt{Strix} can be modified to accommodate local recommendations, or, alternatively, a more sophisticated indicator of a state's global relevance can be learnt.
Second, \texttt{Strix} performs two searches, one for the environment and one for the system player.
However, the parity games we deal with are not entirely symmetric -- environment always moves first.
Thus, we cannot directly apply \texttt{SemML}'s ranking to environment states, as they have a different structure.
Here, we believe that the best solution is to train a separate model for the environment (or rather, six further models).
Thirdly, \texttt{Strix} only constructs the automaton explicitly and computes the game implicitly.
As such, \texttt{Strix} requests scoring information only for edges in the automaton and not in the game.
This can be addressed by closely integrating the scoring computation with the exploration of the automaton -- instead of rebuilding the game for each edge $(p, q)$, we can compute all scores for all outgoing edges of $p$ at once.
Finally, as we mentioned, \texttt{Strix} by default applies a decomposition approach which builds several sub-automata.
These also are equipped with semantic labelling, however with a different meaning -- enough to create a significant hurdle for our learning approach.
We note that \texttt{Strix} actually builds automata by communicating with \texttt{Owl} through a highly optimized interface between Java and C++, significantly complicating passing information back and forth between the processes.

\section{Conclusion}
We demonstrated that semantic labelling can be exploited for practical gains in LTL synthesis.
Our experimental evaluation shows that we vastly outperform the simple approach of \cite{DBLP:conf/atva/KretinskyMM19}, the first step in this direction.
Moreover, despite several mismatches, our approach shows promising results for real world applications of this idea, i.e.\ when combined with the state-of-the-art tool \texttt{Strix}.

\paragraph{Future Work}
As discussed above, the main point for future work is a tight, tailored integration with \texttt{Strix}.
In particular, we want to modify our approach to be applicable to the decomposition methods of \texttt{Strix}, modify \texttt{Strix} to consider local guidance, and actually learn for the precise task required by \texttt{Strix}.

Aside from this, we believe that there might be further interesting features (hand-crafted or learnt) which could provide us with additional insights.
In particular, we want to employ automated feature extraction, through more sophisticated model architectures such as \emph{transformers} or \emph{graph neural networks}.

\clearpage
\bibliographystyle{splncs04}
\bibliography{main}

\begin{thebibliography}{10}
\providecommand{\url}[1]{\texttt{#1}}
\providecommand{\urlprefix}{URL }
\providecommand{\doi}[1]{https://doi.org/#1}

\bibitem{artifact_newest}
{Artifact for "Guessing Winning Policies in LTL Synthesis by Semantic
  Learning"}. Zenodo (Apr 2023). \doi{10.5281/zenodo.7876095}

\bibitem{artifact_eval}
{Artifact for "Guessing Winning Policies in LTL Synthesis by Semantic
  Learning"}. Zenodo (Apr 2023). \doi{10.5281/zenodo.7876096}

\bibitem{DBLP:journals/ita/BernetJW02}
Bernet, J., Janin, D., Walukiewicz, I.: Permissive strategies: from parity
  games to safety games. {RAIRO} Theor. Informatics Appl.  \textbf{36}(3),
  261--275 (2002). \doi{10.1051/ita:2002013}

\bibitem{DBLP:journals/siamcomp/CaludeJKLS22}
Calude, C.S., Jain, S., Khoussainov, B., Li, W., Stephan, F.: Deciding parity
  games in quasi-polynomial time. {SIAM} J. Comput.  \textbf{51}(2),  17--152
  (2022). \doi{10.1137/17m1145288}

\bibitem{DBLP:journals/corr/abs-2303-01158}
Cosler, M., Schmitt, F., Hahn, C., Finkbeiner, B.: Iterative circuit repair
  against formal specifications. CoRR  \textbf{abs/2303.01158} (2023).
  \doi{10.48550/arXiv.2303.01158},
  \url{https://doi.org/10.48550/arXiv.2303.01158}

\bibitem{DBLP:conf/tacas/Dijk18}
van Dijk, T.: Oink: An implementation and evaluation of modern parity game
  solvers. In: Beyer, D., Huisman, M. (eds.) Tools and Algorithms for the
  Construction and Analysis of Systems - 24th International Conference, {TACAS}
  2018, Held as Part of the European Joint Conferences on Theory and Practice
  of Software, {ETAPS} 2018, Thessaloniki, Greece, April 14-20, 2018,
  Proceedings, Part {I}. Lecture Notes in Computer Science, vol. 10805, pp.
  291--308. Springer (2018). \doi{10.1007/978-3-319-89960-2\_16}

\bibitem{DBLP:conf/atva/Duret-LutzLFMRX16}
Duret{-}Lutz, A., Lewkowicz, A., Fauchille, A., Michaud, T., Renault, E., Xu,
  L.: Spot 2.0 - {A} framework for {LTL} and {\textbackslash}omega -automata
  manipulation. In: Artho, C., Legay, A., Peled, D. (eds.) Automated Technology
  for Verification and Analysis - 14th International Symposium, {ATVA} 2016,
  Chiba, Japan, October 17-20, 2016, Proceedings. Lecture Notes in Computer
  Science, vol.~9938, pp. 122--129 (2016). \doi{10.1007/978-3-319-46520-3\_8}

\bibitem{DBLP:conf/cav/EsparzaK14}
Esparza, J., Kret{\'{\i}}nsk{\'{y}}, J.: From {LTL} to deterministic automata:
  {A} safraless compositional approach. In: Biere, A., Bloem, R. (eds.)
  Computer Aided Verification - 26th International Conference, {CAV} 2014, Held
  as Part of the Vienna Summer of Logic, {VSL} 2014, Vienna, Austria, July
  18-22, 2014. Proceedings. Lecture Notes in Computer Science, vol.~8559, pp.
  192--208. Springer (2014). \doi{10.1007/978-3-319-08867-9\_13}

\bibitem{DBLP:conf/tacas/EsparzaKRS17}
Esparza, J., Kret{\'{\i}}nsk{\'{y}}, J., Raskin, J., Sickert, S.: From {LTL}
  and limit-deterministic b{\"{u}}chi automata to deterministic parity
  automata. In: Legay, A., Margaria, T. (eds.) Tools and Algorithms for the
  Construction and Analysis of Systems - 23rd International Conference, {TACAS}
  2017, Held as Part of the European Joint Conferences on Theory and Practice
  of Software, {ETAPS} 2017, Uppsala, Sweden, April 22-29, 2017, Proceedings,
  Part {I}. Lecture Notes in Computer Science, vol. 10205, pp. 426--442 (2017).
  \doi{10.1007/978-3-662-54577-5\_25}

\bibitem{DBLP:journals/sttt/EsparzaKRS22}
Esparza, J., Kret{\'{\i}}nsk{\'{y}}, J., Raskin, J., Sickert, S.: From linear
  temporal logic and limit-deterministic b{\"{u}}chi automata to deterministic
  parity automata. Int. J. Softw. Tools Technol. Transf.  \textbf{24}(4),
  635--659 (2022). \doi{10.1007/s10009-022-00663-1}

\bibitem{DBLP:conf/lics/EsparzaKS18}
Esparza, J., Kret{\'{\i}}nsk{\'{y}}, J., Sickert, S.: One theorem to rule them
  all: {A} unified translation of {LTL} into {\(\omega\)}-automata. In: Dawar,
  A., Gr{\"{a}}del, E. (eds.) Proceedings of the 33rd Annual {ACM/IEEE}
  Symposium on Logic in Computer Science, {LICS} 2018, Oxford, UK, July 09-12,
  2018. pp. 384--393. {ACM} (2018). \doi{10.1145/3209108.3209161}

\bibitem{DBLP:journals/jacm/EsparzaKS20}
Esparza, J., Kret{\'{\i}}nsk{\'{y}}, J., Sickert, S.: A unified translation of
  linear temporal logic to {\(\omega\)}-automata. J. {ACM}  \textbf{67}(6),
  33:1--33:61 (2020). \doi{10.1145/3417995}

\bibitem{DBLP:conf/cav/Fearnley17}
Fearnley, J.: Efficient parallel strategy improvement for parity games. In:
  Majumdar, R., Kuncak, V. (eds.) Computer Aided Verification - 29th
  International Conference, {CAV} 2017, Heidelberg, Germany, July 24-28, 2017,
  Proceedings, Part {II}. Lecture Notes in Computer Science, vol. 10427, pp.
  137--154. Springer (2017). \doi{10.1007/978-3-319-63390-9\_8}

\bibitem{DBLP:conf/spin/FearnleyJS0W17}
Fearnley, J., Jain, S., Schewe, S., Stephan, F., Wojtczak, D.: An ordered
  approach to solving parity games in quasi polynomial time and quasi linear
  space. In: Erdogmus, H., Havelund, K. (eds.) Proceedings of the 24th {ACM}
  {SIGSOFT} International {SPIN} Symposium on Model Checking of Software, Santa
  Barbara, CA, USA, July 10-14, 2017. pp. 112--121. {ACM} (2017).
  \doi{10.1145/3092282.3092286}

\bibitem{DBLP:conf/atva/FriedmannL09}
Friedmann, O., Lange, M.: Solving parity games in practice. In: Liu, Z., Ravn,
  A.P. (eds.) Automated Technology for Verification and Analysis, 7th
  International Symposium, {ATVA} 2009, Macao, China, October 14-16, 2009.
  Proceedings. Lecture Notes in Computer Science, vol.~5799, pp. 182--196.
  Springer (2009). \doi{10.1007/978-3-642-04761-9\_15}

\bibitem{DBLP:conf/atva/GaiserKE12}
Gaiser, A., Kret{\'{\i}}nsk{\'{y}}, J., Esparza, J.: Rabinizer: Small
  deterministic automata for ltl(f, {G)}. In: Chakraborty, S., Mukund, M.
  (eds.) Automated Technology for Verification and Analysis - 10th
  International Symposium, {ATVA} 2012, Thiruvananthapuram, India, October 3-6,
  2012. Proceedings. Lecture Notes in Computer Science, vol.~7561, pp. 72--76.
  Springer (2012). \doi{10.1007/978-3-642-33386-6\_7}

\bibitem{DBLP:conf/atva/HoffmannL13}
Hoffmann, P., Luttenberger, M.: Solving parity games on the {GPU}. In: Hung,
  D.V., Ogawa, M. (eds.) Automated Technology for Verification and Analysis -
  11th International Symposium, {ATVA} 2013, Hanoi, Vietnam, October 15-18,
  2013. Proceedings. Lecture Notes in Computer Science, vol.~8172, pp.
  455--459. Springer (2013). \doi{10.1007/978-3-319-02444-8\_34}

\bibitem{DBLP:journals/corr/abs-2206-00251}
Jacobs, S., P{\'{e}}rez, G.A., Abraham, R., Bruy{\`{e}}re, V., Cadilhac, M.,
  Colange, M., Delfosse, C., van Dijk, T., Duret{-}Lutz, A., Faymonville, P.,
  Finkbeiner, B., Khalimov, A., Klein, F., Luttenberger, M., Meyer, K.J.,
  Michaud, T., Pommellet, A., Renkin, F., Schlehuber{-}Caissier, P., Sakr, M.,
  Sickert, S., Staquet, G., Tamines, C., Tentrup, L., Walker, A.: The reactive
  synthesis competition {(SYNTCOMP):} 2018-2021. CoRR  \textbf{abs/2206.00251}
  (2022). \doi{10.48550/arXiv.2206.00251}

\bibitem{DBLP:journals/ipl/Jurdzinski98}
Jurdzinski, M.: Deciding the winner in parity games is in {UP}
  {\textbackslash}cap co-up. Inf. Process. Lett.  \textbf{68}(3),  119--124
  (1998). \doi{10.1016/S0020-0190(98)00150-1}

\bibitem{DBLP:conf/atva/KomarkovaK14}
Kom{\'{a}}rkov{\'{a}}, Z., Kret{\'{\i}}nsk{\'{y}}, J.: Rabinizer 3: Safraless
  translation of {LTL} to small deterministic automata. In: Cassez, F., Raskin,
  J. (eds.) Automated Technology for Verification and Analysis - 12th
  International Symposium, {ATVA} 2014, Sydney, NSW, Australia, November 3-7,
  2014, Proceedings. Lecture Notes in Computer Science, vol.~8837, pp.
  235--241. Springer (2014). \doi{10.1007/978-3-319-11936-6\_17}

\bibitem{DBLP:conf/atva/KretinskyL13}
Kret{\'{\i}}nsk{\'{y}}, J., Ledesma{-}Garza, R.: Rabinizer 2: Small
  deterministic automata for {LTL} {\(\setminus\)} {GU}. In: Hung, D.V., Ogawa,
  M. (eds.) Automated Technology for Verification and Analysis - 11th
  International Symposium, {ATVA} 2013, Hanoi, Vietnam, October 15-18, 2013.
  Proceedings. Lecture Notes in Computer Science, vol.~8172, pp. 446--450.
  Springer (2013). \doi{10.1007/978-3-319-02444-8\_32}

\bibitem{DBLP:conf/atva/KretinskyMM19}
Kret{\'{\i}}nsk{\'{y}}, J., Manta, A., Meggendorfer, T.: Semantic labelling and
  learning for parity game solving in {LTL} synthesis. In: Chen, Y., Cheng, C.,
  Esparza, J. (eds.) Automated Technology for Verification and Analysis - 17th
  International Symposium, {ATVA} 2019, Taipei, Taiwan, October 28-31, 2019,
  Proceedings. Lecture Notes in Computer Science, vol. 11781, pp. 404--422.
  Springer (2019). \doi{10.1007/978-3-030-31784-3\_24}

\bibitem{DBLP:conf/atva/KretinskyMS18}
Kret{\'{\i}}nsk{\'{y}}, J., Meggendorfer, T., Sickert, S.: Owl: {A} library for
  {\(\omega\)}-words, automata, and {LTL}. In: Lahiri, S.K., Wang, C. (eds.)
  Automated Technology for Verification and Analysis - 16th International
  Symposium, {ATVA} 2018, Los Angeles, CA, USA, October 7-10, 2018,
  Proceedings. Lecture Notes in Computer Science, vol. 11138, pp. 543--550.
  Springer (2018). \doi{10.1007/978-3-030-01090-4\_34}

\bibitem{DBLP:conf/cav/KretinskyMSZ18}
Kret{\'{\i}}nsk{\'{y}}, J., Meggendorfer, T., Sickert, S., Ziegler, C.:
  Rabinizer 4: From {LTL} to your favourite deterministic automaton. In:
  Chockler, H., Weissenbacher, G. (eds.) Computer Aided Verification - 30th
  International Conference, {CAV} 2018, Held as Part of the Federated Logic
  Conference, FloC 2018, Oxford, UK, July 14-17, 2018, Proceedings, Part {I}.
  Lecture Notes in Computer Science, vol. 10981, pp. 567--577. Springer (2018).
  \doi{10.1007/978-3-319-96145-3\_30}

\bibitem{DBLP:journals/acta/KretinskyMWW22}
Kret{\'{\i}}nsk{\'{y}}, J., Meggendorfer, T., Waldmann, C., Weininger, M.:
  Index appearance record with preorders. Acta Informatica  \textbf{59}(5),
  585--618 (2022). \doi{10.1007/s00236-021-00412-y},
  \url{https://doi.org/10.1007/s00236-021-00412-y}

\bibitem{DBLP:conf/mochart/KupfermanR10}
Kupferman, O., Rosenberg, A.: The blowup in translating {LTL} to deterministic
  automata. In: van~der Meyden, R., Smaus, J. (eds.) Model Checking and
  Artificial Intelligence - 6th International Workshop, MoChArt 2010, Atlanta,
  GA, USA, July 11, 2010, Revised Selected and Invited Papers. Lecture Notes in
  Computer Science, vol.~6572, pp. 85--94. Springer (2010).
  \doi{10.1007/978-3-642-20674-0\_6}

\bibitem{DBLP:journals/lmcs/LehtinenPSW22}
Lehtinen, K., Parys, P., Schewe, S., Wojtczak, D.: A recursive approach to
  solving parity games in quasipolynomial time. Log. Methods Comput. Sci.
  \textbf{18}(1) (2022). \doi{10.46298/lmcs-18(1:8)2022}

\bibitem{10.1109/TIME.2013.19}
Li, J., Zhang, L., Pu, G., Vardi, M.Y., He, J.: {LTL} satisfiability checking
  revisited. In: S{\'{a}}nchez, C., Venable, K.B., Zim{\'{a}}nyi, E. (eds.)
  2013 20th International Symposium on Temporal Representation and Reasoning,
  Pensacola, FL, USA, September 26-28, 2013. pp. 91--98. {IEEE} Computer
  Society (2013). \doi{10.1109/TIME.2013.19}

\bibitem{liu_09_LTR}
Liu, T.: Learning to rank for information retrieval. Found. Trends Inf. Retr.
  \textbf{3}(3),  225--331 (2009). \doi{10.1561/1500000016}

\bibitem{DBLP:journals/acta/LuttenbergerMS20}
Luttenberger, M., Meyer, P.J., Sickert, S.: Practical synthesis of reactive
  systems from {LTL} specifications via parity games. Acta Informatica
  \textbf{57}(1-2),  3--36 (2020). \doi{10.1007/s00236-019-00349-3}

\bibitem{DBLP:conf/atva/MeyerL16}
Meyer, P.J., Luttenberger, M.: Solving mean-payoff games on the {GPU}. In:
  Artho, C., Legay, A., Peled, D. (eds.) Automated Technology for Verification
  and Analysis - 14th International Symposium, {ATVA} 2016, Chiba, Japan,
  October 17-20, 2016, Proceedings. Lecture Notes in Computer Science,
  vol.~9938, pp. 262--267 (2016). \doi{10.1007/978-3-319-46520-3\_17}

\bibitem{DBLP:conf/cav/MeyerSL18}
Meyer, P.J., Sickert, S., Luttenberger, M.: Strix: Explicit reactive synthesis
  strikes back! In: Chockler, H., Weissenbacher, G. (eds.) Computer Aided
  Verification - 30th International Conference, {CAV} 2018, Held as Part of the
  Federated Logic Conference, FloC 2018, Oxford, UK, July 14-17, 2018,
  Proceedings, Part {I}. Lecture Notes in Computer Science, vol. 10981, pp.
  578--586. Springer (2018). \doi{10.1007/978-3-319-96145-3\_31}

\bibitem{osborne2004introduction}
Osborne, M.J.: An introduction to game theory  (2004)

\bibitem{scikit-learn}
Pedregosa, F., Varoquaux, G., Gramfort, A., Michel, V., Thirion, B., Grisel,
  O., Blondel, M., Prettenhofer, P., Weiss, R., Dubourg, V., Vanderplas, J.,
  Passos, A., Cournapeau, D., Brucher, M., Perrot, M., Duchesnay, E.:
  Scikit-learn: Machine learning in {P}ython. Journal of Machine Learning
  Research  \textbf{12},  2825--2830 (2011)

\bibitem{DBLP:conf/lics/Piterman06}
Piterman, N.: From nondeterministic buchi and streett automata to deterministic
  parity automata. In: 21th {IEEE} Symposium on Logic in Computer Science
  {(LICS} 2006), 12-15 August 2006, Seattle, WA, USA, Proceedings. pp.
  255--264. {IEEE} Computer Society (2006). \doi{10.1109/LICS.2006.28}

\bibitem{Pnueli77}
Pnueli, A.: The temporal logic of programs. In: 18th Annual Symposium on
  Foundations of Computer Science, Providence, Rhode Island, USA, 31 October -
  1 November 1977. pp. 46--57. {IEEE} Computer Society (1977).
  \doi{10.1109/SFCS.1977.32}

\bibitem{DBLP:conf/icalp/PnueliR89}
Pnueli, A., Rosner, R.: On the synthesis of an asynchronous reactive module.
  In: Ausiello, G., Dezani{-}Ciancaglini, M., Rocca, S.R.D. (eds.) Automata,
  Languages and Programming, 16th International Colloquium, ICALP89, Stresa,
  Italy, July 11-15, 1989, Proceedings. Lecture Notes in Computer Science,
  vol.~372, pp. 652--671. Springer (1989). \doi{10.1007/BFb0035790}

\bibitem{DBLP:conf/focs/Safra88}
Safra, S.: On the complexity of omega-automata. In: 29th Annual Symposium on
  Foundations of Computer Science, White Plains, New York, USA, 24-26 October
  1988. pp. 319--327. {IEEE} Computer Society (1988).
  \doi{10.1109/SFCS.1988.21948}

\bibitem{DBLP:conf/fossacs/Schewe09}
Schewe, S.: Tighter bounds for the determinisation of b{\"{u}}chi automata. In:
  de~Alfaro, L. (ed.) Foundations of Software Science and Computational
  Structures, 12th International Conference, {FOSSACS} 2009, Held as Part of
  the Joint European Conferences on Theory and Practice of Software, {ETAPS}
  2009, York, UK, March 22-29, 2009. Proceedings. Lecture Notes in Computer
  Science, vol.~5504, pp. 167--181. Springer (2009).
  \doi{10.1007/978-3-642-00596-1\_13}

\bibitem{DBLP:conf/nips/SchmittHRF21}
Schmitt, F., Hahn, C., Rabe, M.N., Finkbeiner, B.: Neural circuit synthesis
  from specification patterns. In: Ranzato, M., Beygelzimer, A., Dauphin, Y.N.,
  Liang, P., Vaughan, J.W. (eds.) Advances in Neural Information Processing
  Systems 34: Annual Conference on Neural Information Processing Systems 2021,
  NeurIPS 2021, December 6-14, 2021, virtual. pp. 15408--15420 (2021),
  \url{https://proceedings.neurips.cc/paper/2021/hash/8230bea7d54bcdf99cdfe85cb07313d5-Abstract.html}

\bibitem{Sickert_16_LDBA}
Sickert, S., Esparza, J., Jaax, S., K{\v{r}}et{\'{i}}nsk{\'{y}}, J.:
  Limit-deterministic b{\"{u}}chi automata for linear temporal logic. In:
  Chaudhuri, S., Farzan, A. (eds.) Computer Aided Verification - 28th
  International Conference, {CAV} 2016, Toronto, ON, Canada, July 17-23, 2016,
  Proceedings, Part {II}. Lecture Notes in Computer Science, vol.~9780, pp.
  312--332. Springer (2016). \doi{10.1007/978-3-319-41540-6\_17}

\bibitem{DBLP:conf/lics/VardiW86}
Vardi, M.Y., Wolper, P.: An automata-theoretic approach to automatic program
  verification (preliminary report). In: Proceedings of the Symposium on Logic
  in Computer Science {(LICS} '86), Cambridge, Massachusetts, USA, June 16-18,
  1986. pp. 332--344. {IEEE} Computer Society (1986)

\end{thebibliography}

\iftoggle{arxiv}{
\newpage
\appendix
%\section{Further data} \label{sec:appendix:data}
%
%
\section{Further Details} \label{sec:appendix:details}

%\subsection{SI bad} \label{sec:appendix:si_bad}
%
%\todo{blub}
%
%\begin{figure}[t]
%	\centering
%	\begin{tikzpicture}[auto,yscale=0.8]
%		\node[systemltl, initial left] (1) at (0,0) {$v_1,1$};
%		\node[systemltl] (2) at (3,1) {$v_2,2$};
%		\node[systemltl] (3) at (3,-1) {$v_3,2$};
%		\node[systemltl] (T) at (6,0) {$v_\top,1$};
%		
%		\path[->]
%			(1) edge[bend left=15] (2)
%			(1) edge[bend right=15,swap] (3)
%			(2) edge[bend right=20,swap] (3)
%			(3) edge[bend right=20,swap] (2)
%			(2) edge[loop above] (2)
%			(3) edge[loop below] (3)
%			(2) edge[bend left=15] (T)
%			(3) edge[bend right=15,swap] (T)
%			(T) edge[loop above] (T)
%		;
%	\end{tikzpicture}
%	\caption{
%		Simple parity game where SI cannot predict a sensible ground truth.
%		}\label{fig:SI_bad_extended}
%\end{figure}

\subsection{A Simple Realistic Automaton State} \label{sec:appendix:realistic_example}

To further illustrate the complexity of the semantic labelling in real-world applications, we present another state pair obtained from \texttt{Owl} in \cref{fig:example_labelling_real}, this time for the formula $(\ltlGlobally \ltlFinally(\mathrm{p} \Leftrightarrow \ltlNext \ltlNext \mathrm{q})) \Leftrightarrow (\ltlGlobally \ltlFinally \mathrm{acc})$, one of the smallest SYNTCOMP formulae, called \texttt{ltl2dba17}.
The resulting automaton has 50 such states and 400 transitions.

Another example, obtained from a small random formula, is given in \cref{fig:example_labelling_extra_complicated}.
%formula: $((a \land \ltlGlobally b) \lor (\ltlGlobally \ltlFinally(c) \land \ltlGlobally((\neg a | \neg b) \ltlUntil (\ltlNext c))))$

\begin{figure}[p]
	\centering
	\begin{tikzpicture}[every node/.style={scale=0.8}]
	\node[rectangle,draw,align=center] (q0) at (0,0) {%
		\begin{minipage}{\textwidth}
		\vspace{-1em}
		\begin{gather*}
			\big(((\neg acc \land \ltlGlobally \neg acc) \lor \ltlFinally \ltlGlobally \neg acc) \land (((p \lor \ltlNext \ltlNext q) \land \ltlGlobally (p \lor \ltlNext \ltlNext q)) \lor \ltlFinally \ltlGlobally (p \lor \ltlNext \ltlNext q)) \land {} \\
			(((\neg p \lor \ltlNext \ltlNext \neg q) \land \ltlGlobally (\neg p \lor \ltlNext \ltlNext \neg q)) \lor \ltlFinally \ltlGlobally (\neg p \lor \ltlNext \ltlNext \neg q))\big) \lor {} \\
			\big((acc \lor \ltlFinally acc) \land \ltlGlobally \ltlFinally acc \land ((((p \land \ltlNext \ltlNext q) \lor \ltlFinally (p \land \ltlNext \ltlNext q)) \land \ltlGlobally \ltlFinally (p \land \ltlNext \ltlNext q)) \lor {} \\
			(((\neg p \land \ltlNext \ltlNext \neg q) \lor \ltlFinally (\neg p \land \ltlNext \ltlNext \neg q)) \land \ltlGlobally \ltlFinally (\neg p \land \ltlNext \ltlNext \neg q)))\big)
		\end{gather*}
		\end{minipage} \\
		\begin{tabular}{lcc}
			\midrule
			\multirow{3}{*}{$M_1$:} &  co-safety:  &                                                                                                   $[(acc \lor \ltlFinally acc)]$                                                                                                   \\
			                        & f-co-safety: &                                                           $[(acc \lor \ltlFinally acc), ((p \land \ltlNext \ltlNext q) \lor \ltlFinally (p \land \ltlNext \ltlNext q))]$                                                           \\
			                        &   safety:    &                                                                                                              $\true$                                                                                                               \\
			\midrule
			\multirow{3}{*}{$M_2$:} &  co-safety:  &                                                                                                   $[(acc \lor \ltlFinally acc)]$                                                                                                   \\
			                        & f-co-safety: &                                                 $[(acc \lor \ltlFinally acc), ((\neg p \land \ltlNext \ltlNext \neg q) \lor \ltlFinally (\neg p \land \ltlNext \ltlNext \neg q))]$                                                 \\
			                        &   safety:    &                                                                                                              $\true$                                                                                                               \\
			\midrule
			\multirow{3}{*}{$M_3$:} &  co-safety:  &                                                                                                             $[\true]$                                                                                                              \\
			                        & f-co-safety: &                                                                                                                $[]$                                                                                                                \\
			                        &   safety:    & $(\neg acc \land \ltlGlobally \neg acc \land (p \lor \ltlNext \ltlNext q) \land (\neg p \lor \ltlNext \ltlNext \neg q) \land \ltlGlobally (p \lor \ltlNext \ltlNext q) \land \ltlGlobally (\neg p \lor \ltlNext \ltlNext \neg q))$
		\end{tabular}
	};

	\node[rectangle,draw,align=center] (q1) at (0,-6.5) {%
		\begin{minipage}{\textwidth}
		\vspace{-1em}
		\begin{gather*}
			\big(((\neg acc \land \ltlGlobally \neg acc) \lor \ltlFinally \ltlGlobally \neg acc) \land (((p \lor \ltlNext \ltlNext q) \land \ltlGlobally (p \lor \ltlNext \ltlNext q)) \lor \ltlFinally \ltlGlobally (p \lor \ltlNext \ltlNext q)) \land {} \\
			(((\neg p \lor \ltlNext \ltlNext \neg q) \land \ltlGlobally (\neg p \lor \ltlNext \ltlNext \neg q)) \lor \ltlFinally \ltlGlobally (\neg p \lor \ltlNext \ltlNext \neg q))\big) \lor {} \\
			\big((acc \lor \ltlFinally acc) \land \ltlGlobally \ltlFinally acc \land ((((p \land \ltlNext \ltlNext q) \lor \ltlFinally (p \land \ltlNext \ltlNext q)) \land \ltlGlobally \ltlFinally (p \land \ltlNext \ltlNext q)) \lor {} \\
			(((\neg p \land \ltlNext \ltlNext \neg q) \lor \ltlFinally (\neg p \land \ltlNext \ltlNext \neg q)) \land \ltlGlobally \ltlFinally (\neg p \land \ltlNext \ltlNext \neg q)))\big)
		\end{gather*}
		\end{minipage} \\
		\begin{tabular}{lcc}
			\midrule
			\multirow{3}{*}{$M_1$:} &  co-safety:  &                                                                         $[((p \land \ltlNext \ltlNext q) \lor \ltlFinally (p \land \ltlNext \ltlNext q))]$                                                                         \\
			                                             & f-co-safety: &                                                           $[(acc \lor \ltlFinally acc), ((p \land \ltlNext \ltlNext q) \lor \ltlFinally (p \land \ltlNext \ltlNext q))]$                                                           \\
			                                             &   safety:    &                                                                                                              $\true$                                                                                                               \\
			\midrule
			\multirow{3}{*}{$M_2$:}                      &  co-safety:  &                                                               $[((\neg p \land \ltlNext \ltlNext \neg q) \lor \ltlFinally (\neg p \land \ltlNext \ltlNext \neg q))]$                                                               \\
			                                             & f-co-safety: &                                                 $[(acc \lor \ltlFinally acc), ((\neg p \land \ltlNext \ltlNext \neg q) \lor \ltlFinally (\neg p \land \ltlNext \ltlNext \neg q))]$                                                 \\
			                                             &   safety:    &                                                                                                              $\true$                                                                                                               \\
			\midrule
			\multirow{3}{*}{$M_3$:}                      &  co-safety:  &                                                                                                             $[\true]$                                                                                                              \\
			                                             & f-co-safety: &                                                                                                                $[]$                                                                                                                \\
			                                             &   safety:    & $(\neg acc \land \ltlGlobally \neg acc \land (p \lor \ltlNext \ltlNext q) \land (\neg p \lor \ltlNext \ltlNext \neg q) \land \ltlGlobally (p \lor \ltlNext \ltlNext q) \land \ltlGlobally (\neg p \lor \ltlNext \ltlNext \neg q))$
		\end{tabular}
	};
	\path[thick,->] (q0) edge node[right] {$\{p \mapsto \true, q \mapsto \true, acc \mapsto \true\}$, $4$} (q1);
	\end{tikzpicture}
	\caption{
		A single transition in the automaton for $(\ltlGlobally \ltlFinally(p \Leftrightarrow \ltlNext \ltlNext q)) \Leftrightarrow (\ltlGlobally \ltlFinally acc)$.
	} \label{fig:example_labelling_real}
\end{figure}
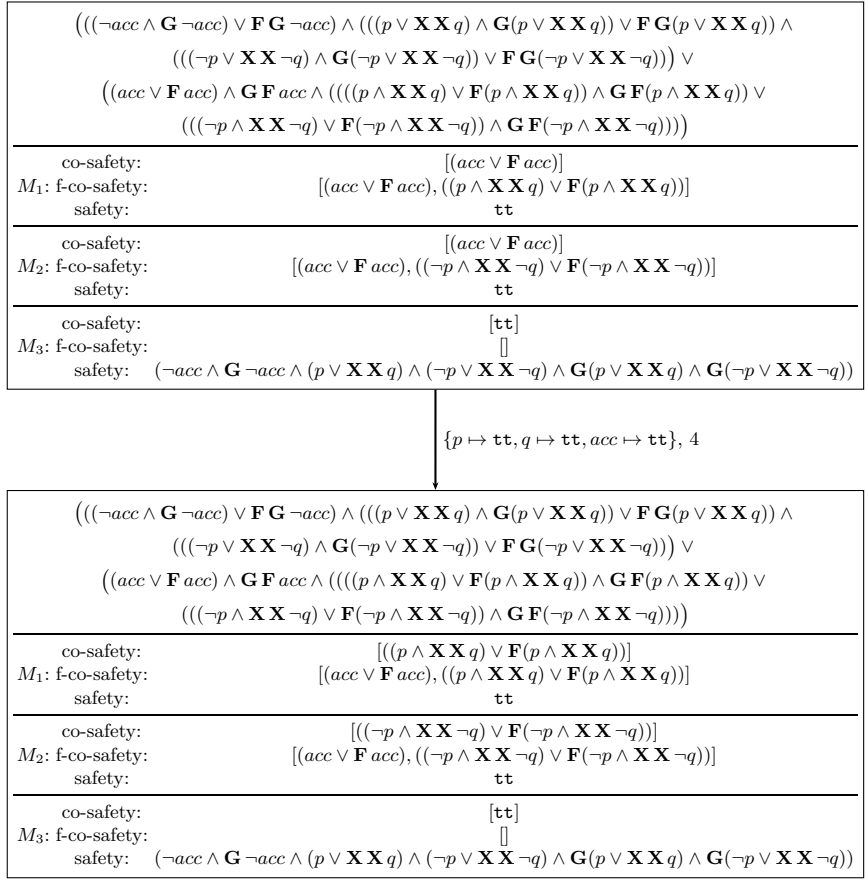

\begin{figure}[t]
	\centering
	\begin{tikzpicture}[every node/.style={scale=0.8}]
	\node[rectangle,draw] (q0) at (0,0) {
		\begin{tabular}{lcc}
			\multicolumn{3}{c}{ $((a \land b \land \ltlGlobally b) \lor ((c\lor \ltlFinally c) \land (\neg a \lor \neg b \lor \ltlNext c) \land \ltlGlobally\ltlFinally c \land \ltlGlobally(\neg a\lor \neg b \ltlNext c)))$} \\
			\midrule
			\multirow{2}{*}{$M_1$:} & co-safety: &                                                                          $[c \lor \ltlFinally c]$                                                                           \\
			                        &  safety:   &                                       $((\neg a \lor \neg b \lor \ltlNext c)\land \ltlGlobally (\neg a \lor \neg b \lor \ltlNext c))$                                       \\
			\midrule
			\multirow{2}{*}{$M_2$:} & co-safety: &                                                                                  $[\true]$                                                                                  \\
			                        &  safety:   &                                                                      $a \land b \land \ltlGlobally b$
		\end{tabular}
	};

	\node[rectangle,draw] (q1) at (0,-4) {
		\begin{tabular}{lcc}
			\multicolumn{3}{c}{$((b \land \ltlGlobally b) \lor (c \land (c \lor \ltlFinally c) \land (\neg a \lor \neg b \lor \ltlNext c) \land \ltlGlobally \ltlFinally c \land \ltlGlobally(\neg a \lor \neg b \lor \ltlNext c))) $} \\
			\midrule
			\multirow{2}{*}{$M_1$:} & co-safety: &                                                                                      $[\true]$                                                                                      \\
			                        &  safety:   &                        $(c \land (c \lor \ltlFinally c) \land (\neg a \lor \neg b \lor \ltlNext c) \land \ltlGlobally(\neg a \lor \neg b \lor \ltlNext c)) $                        \\
			\midrule
			\multirow{2}{*}{$M_2$:} & co-safety: &                                                                                      $[\true]$                                                                                      \\
			                        &  safety:   &                                                                              $b \land \ltlGlobally b$
		\end{tabular}
	};
	
	\path[thick,->] (q0) edge node[right] {$\{a \mapsto \true, b \mapsto \true, c \mapsto \false\}$} (q1);
	\end{tikzpicture}
	\caption{
		A single transition in the automaton for $(a \land \ltlGlobally b) \lor (\ltlGlobally \ltlFinally c \land \ltlGlobally ((\neg a \lor \neg b) \ltlUntil (\ltlNext c)))$.
	} \label{fig:example_labelling_extra_complicated}
\end{figure}

\subsection{Definition of Obligation Sets} \label{app:osf_def}

We inductively define the obligation set formula $\obligationSetFormula$ as follows:
\newcommand{\osf}{\obligationSetFormula}
\begin{align*}
	\osf(\true)                            & = \true                                 \\
	\osf(\false)                           & = \false                                \\
	\osf(a)                                & = a \text{ for } a\in\AP                \\
	\osf(\neg a)                           & = \neg a  \text{ for } a\in\AP          \\
	\osf(\varphi_1 \land \varphi_2)        & = \osf(\varphi_1) \land \osf(\varphi_2) \\
	\osf(\varphi_1 \lor \varphi_2)         & = \osf(\varphi_1) \lor \osf(\varphi_2)  \\
	\osf(\ltlNext \varphi_1)               & = \osf(\varphi_1)                       \\
	\osf(\ltlFinally \varphi_1)            & = \osf(\varphi_1)                       \\
	\osf(\ltlGlobally \varphi_1)           & = \osf(\varphi_1)                       \\
	\osf(\varphi_1 \ltlUntil \varphi_2)    & = \osf(\varphi_2)                       \\
	\osf(\varphi_1 \ltlRelease \varphi_2)  & = \osf(\varphi_2)                       \\
	\osf(\varphi_1 \ltlWuntil \varphi_2)   & = \osf(\varphi_1) \lor \osf(\varphi_2)  \\
	\osf(\varphi_1 \ltlSrelease \varphi_2) & = \osf(\varphi_1) \land \osf(\varphi_2)
\end{align*}}
{}.
%% Appendix

\end{document}